\documentclass[11pt]{article}
\usepackage{basicarticle}
\usepackage{float}
\usepackage{placeins}

\papertitle{Measuring the evolution of camera distance across a century of film}

\paperauthors{%
  David Bamman\orcidlink{0009-0003-1171-9408}$^{1,\ast}$ \quad
  Allison Cooper\orcidlink{0000-0002-5738-3562}$^{2}$ \quad
  Dan Hickey\orcidlink{0009-0006-5577-0063}$^{1}$ \quad
  Madison Mar\orcidlink{0009-0002-8252-3755}$^{1}$}

\paperaffiliations{%
  $^{1}$School of Information, University of California, Berkeley, CA 94704, USA
  \\$^{2}$Cinema Studies Program, Bowdoin College, Brunswick, ME 04011, USA}

\papercorresp{$^\ast$Correspondence: \href{mailto:dbamman@berkeley.edu}{dbamman@berkeley.edu}}

\paperabstract{The rise of computer vision and artificial intelligence has made possible new forms of large-scale computational measurement. We apply these techniques to a deep collection of 5,205 digitized films viewed in theaters between 1922-2025 (covering popular, prestigious, and independent movies) to trace the development of one of the most fundamental ways through which film communicates: by manipulating the space between the camera and its subject.  This work finds abrupt changes with the rise of new technologies in sound and television, and allows us to shed an empirical light on gender disparity (women, despite having substantially less screentime than men, are disproportionately the subject of closer shots), and illustrate how animated films both inherit and break free from the norms of live-action filmmaking.}

\paperkeywords{Film $\mid$ Computer vision $\mid$ AI $\mid$ Culture}

\begin{document}
\maketitleblock

Onscreen storytelling has increasingly come to mediate our experience of the world. Since its recordkeeping began in 2003, the American Time Use Survey has regularly found that Americans spend roughly 10 times as long watching TV as reading for pleasure\ \cite{bls_atus_2024}, and the total number of feature films released per year has moved from a roughly linear trend since the birth of narrative film  to an  exponential increase at the turn of the millennium (fig. \ref{fig:imdbfeatures}, SI).

Despite its growing significance, what we know about the formal storytelling strategies of narrative cinema is largely the result of modestly scoped, qualitative studies undertaken by film and media studies researchers. This was the case, for example, in defining ``Classical Hollywood'' cinema, a style of filmmaking favored by the studio system during the Golden Age of Hollywood. By closely viewing 100 Hollywood films produced between 1915 and 1960 and related industrial documents, David Bordwell et al. \cite{bordwell1985classical} identified a recurring narrative pattern of psychologically individuated characters who overcome obstacles to achieve their objectives, undergoing significant changes in situation, attitude, or values along the way. Bordwell et al.'s study also demonstrated the extent to which narrative film’s formal strategies coalesced in the Classical Hollywood era to create coherent storytelling onscreen through the continuity system, an editing strategy designed to ensure a seamless viewing experience for the spectator. Remarkably, however, the 100 films that made up their corpus comprised just 0.3\% of the 29,998 feature films released in the U.S. over their period of study \cite[388]{bordwell1985classical}.  Film historian Barry Salt offers an empirical complement to this work; his work on statistical style analysis\ \cite{d0d389e1-a831-31fa-95b0-d6e44511b8e9} opened the door to the quantitative study of film, focusing on empirical measurements of shot length, camera movement and shot scale; but while his most voluminous work\ \cite{salt2009film} measuring average shot length spans close to 10,000 films, his work on camera distance is more modestly scoped to approximately 200 movies.

Advances in machine learning and computer vision, however---along with new access to film data in the United States---have recently made it possible to measure cinema’s narrative and formal characteristics across a substantially larger corpus that reflects the global expansion of digital media into the 21st century \cite{manovich2012compare}.  
In this work, we exploit these developments to digitize a 5,205-film dataset comprised of prestige, popular, and independent films dating from 1922 through the present, in order to investigate the changing dynamics of a fundamental way in which film communicates: by varying the relationship between the camera and its subject. 

The manipulation of space is a key element of onscreen storytelling, carried out at the level of mise-en-scene, where setting, lighting, and staging help create the illusion of depth and surface; cinematography, where lens choice, framing, and camera movement determine static and dynamic spatial relations; and editing, where the cut can create or disrupt spatial coherence. Training manuals for practitioners---directors, editors, and cinematographers---offer time-honored, detailed instructions on the production of space onscreen, situating it within the overall grammar of film \cite{arijon1976grammar,mascelli1965five,reisz1972technique,bowen2013grammar,block2013visual}. Academic accounts parallel industry accounts, theorizing cinematic space at a more abstract level, from classical film theory’s arguments in favor of spatial realism \cite{bazin2004cinema,kracauer1997theory}  to structuralist approaches that analyze the function of narrative space within broader systems \cite{heath1976narrative,burch2014theory}. Cognitive film theory, on the other hand, offers a counterargument to psychoanalytic and ideological interpretations of cinematic space by focusing on the mental models mustered by the spectator to create a coherent spatial representation out of what is presented onscreen \cite{bordwell1985narration}. For cognitivists, camera distance offers the viewer scalar information to build mental models, from the large-scale spatial relationships provided by the extreme long shot to the emotion or detail distilled in the more constrained views provided by the close-up. Cognitivist theory also draws on studies of how humans experience interpersonal interactions \cite{hall1990hidden} to analyze the impact of camera distance on the spectator, arguing that the mechanisms that regulate in-person encounters also inform the viewer’s response to something like the close-up onscreen \cite{plantinga2009moving}.

Industrial accounts of narrative space focus primarily on how creators can construct a spatially coherent story world for the spectator, whereas theoretical accounts focus on the ways narrative space carries meaning, produces affect, or engages cognitive processes. Our study draws upon both traditions as we operationalize industrial definitions of camera distance to conduct computational, corpus-level research that can both respond to and raise new theoretical questions about the role it plays in onscreen storytelling.  

Our findings speak to three sets of research questions.  First, in measuring how the apparent distance between the camera and its subject evolved over the the past 100 years of filmmaking, we expect changes to center around major technological innovations with the introduction of sound, the rise of television in the 1950s and smartphone viewing in the 2010s; we confirm two of these changes in our data (sound and television).  Second, in examining whether camera distance is a gendered phenomenon, we draw on work in feminist film theory---including Mulvey's work on the male gaze\ \cite{Mulvey1975}---to measure the degree to which women are the subject of medium close-ups more often than men; we find that while women are disproportionately underrepresented on screen, they appear in medium close-ups 20\% more frequently than men.  And finally, in asking how animation---freed from the constraints of a physical camera and physical space---uses camera distance in its own storytelling, we find that it uses a much longer shot distances than live-action film, and does not exhibit the same gender differences in the medium close-up for either human characters or non-human ones, providing important context to our finding on live-action gender representation.

\section{Data}

We draw our collection of films from three separate sources: popular (based on historical box office earnings), prestige (based on award nominations), and independent (based on scholarly inclusion criteria).  Each subcollection offers a different vantage point on the use of camera distance.

\subsection{Popular}

We leverage two sources of information for popular movies, both establishing popularity through box office revenues earned each calendar year.  For the period 1980-2025, we draw on \url{https://www.boxofficemojo.com/} (Domestic Yearly Box Office calendar grosses); for the period 1922-1979, we draw on data released by\ Bamman et al. 2026 \cite{bamman2026evaluating}, which extracts weekly box office information reported by \emph{Variety} magazine for major metropolitan areas in the United States.  We select up to the top 50 movies by box office for each year, including both live-action and animated films, yielding a total of 4,064 unique movies.

\subsection{Prestige}

To capture a marker of prestige, we draw on several diverse sources: first, award nominations from 12 different organizations (Academy Awards, BAFTA, Berlin Film Festival, Cannes Film Festival, Golden Globes, LAFCA, National Society of Film Critics, NBR, Sundance, Toronto International Film Festival, Tribeca Prize, Venice Film Festival), largely focusing on major film and filmmaking awards (as opposed to performance awards), such as Best Picture, Best Cinematography, and the Cannes Palme d'Or. For each year, we rank films by the total number of award nominations they received among these organizations and prioritized acquisition of the most awarded. We additionally include movies listed in several prestigious rankings, including the AFI Top 100 Movies, \emph{The New York Times} 100 Best Movies of the 21st Century, the BFI Greatest Films of All Time, and the BFI Directors’ 100 Greatest Films of All Time, yielding a total of 1,883 unique movies.  

\subsection{American Independent}

Finally, in order to capture an alternative point of view that does not directly align with either a popular or prestigious axis, we selected a sample of American independent films listed in the \emph{Directory of World Cinema: American Independent} (volumes 1 and 2) edited by John Berra\ \cite{berra2010directory,berra2013american}, yielding a total of 226 movies.
\\[10pt]
\noindent
We purchase each movie and digitize it under US 37 CFR 201.40(b)(4), which permits breaking technological protection measures on DVDs and Blu-ray disks for text and data mining research.  In total, the collection spans 5,205 unique movies viewed in theaters over the period 1922-2025.  Not all films are able to be acquired over our period of study; figs \ref{fig:popular}-\ref{fig:indie} (SI) show the distribution for each subcollection over time. While our selection criterion prioritizes contemporaneous interest with the film's release (e.g., through either historical box office popularity or award nominations), data availability must also reflect later popular and commercial interests as well.

\begin{figure*}[t!]
\centering

\begin{subfigure}[t]{0.325\linewidth}
    \centering
    \includegraphics[width=\linewidth,height=3.8cm,keepaspectratio]{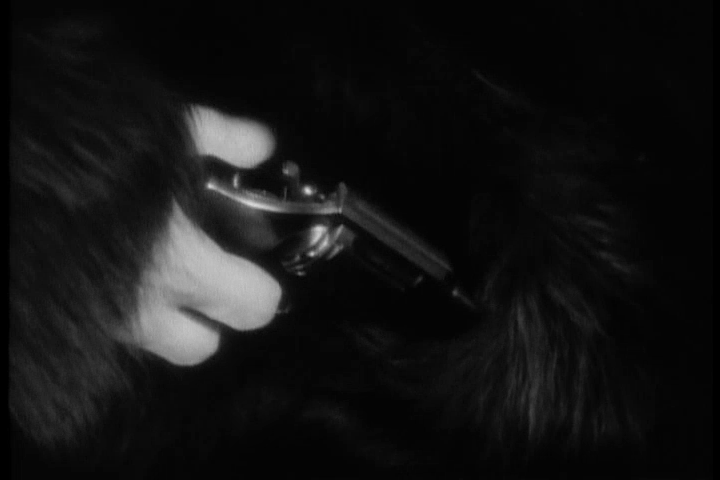}
    \caption{Close-up}
\end{subfigure}
\hfill
\begin{subfigure}[t]{0.325\linewidth}
    \centering
    \includegraphics[width=\linewidth,height=3.8cm,keepaspectratio]{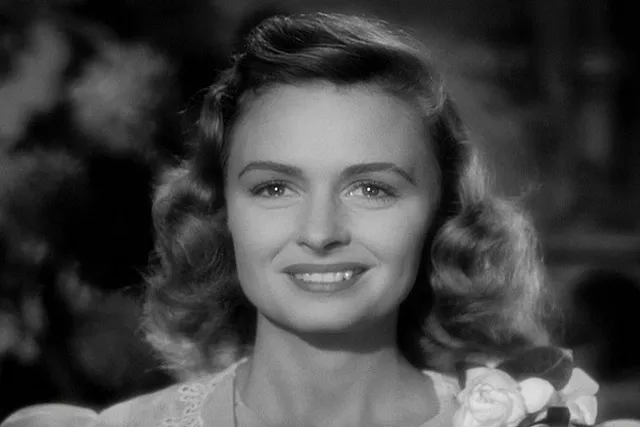}
    \caption{Medium close-up}
\end{subfigure}
\hfill
\begin{subfigure}[t]{0.325\linewidth}
    \centering
    \includegraphics[width=\linewidth,height=3.8cm,keepaspectratio]{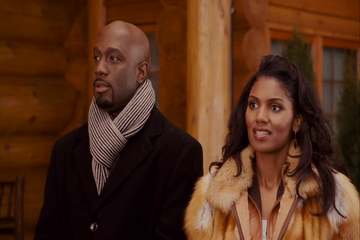}
    \caption{Medium}
\end{subfigure}

\vspace{0.5cm}

\begin{subfigure}[t]{0.325\linewidth}
    \centering
    \includegraphics[width=\linewidth,height=3.8cm,keepaspectratio]{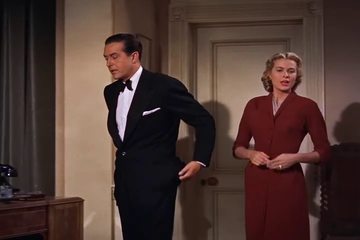}
    \caption{Medium long}
\end{subfigure}
\hfill
\begin{subfigure}[t]{0.325\linewidth}
    \centering
    \includegraphics[width=\linewidth,height=3.8cm,keepaspectratio]{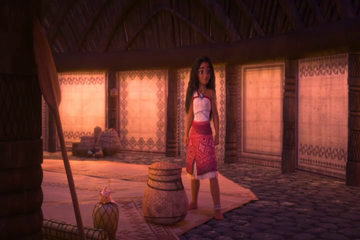}
    \caption{Long}
\end{subfigure}
\hfill
\begin{subfigure}[t]{0.325\linewidth}
    \centering
    \includegraphics[width=\linewidth,height=3.8cm,keepaspectratio]{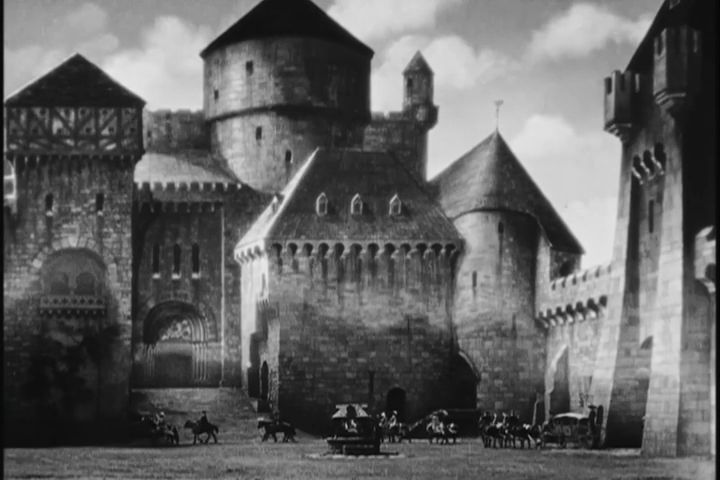}
    \caption{Extreme long}
\end{subfigure}

\caption{Examples of distance categories.}
\label{fig:distanceexamples}
\end{figure*}
\section{Measuring distance}

Space is a key aspect of \emph{story}, the intentionally arranged presentation of narrative elements that differentiate it from the raw \textit{fabula}, the linear chronology in which those events took place in the story's world \cite{bal2017narratology}. 
Space’s role in film narrative is significant since, in addition to producing the visual world that is constitutive of cinema, it can subjectivize the fabula’s geographical locations or physical settings by representing them through the filter of perception---that is, by linking them to a process of focalization that might belong to a character or to the narration itself. For one significant school of thought---screen theory---camera distances and associated shot sizes are effectively part of how the film apparatus can constitute spectators by positioning, interpellating, and subjecting them ideologically and psychically \cite{baudry1974ideological,heath1976narrative}. Cognitivist film theory, on the other hand, argues that camera distance plays an important role in positioning an active, meaning-making viewer in relation to characters’ perceptual or psychological experiences \cite{branigan2013narrative}---even postulating that a shot like the close-up engages the viewer’s motor and affective systems to produce somatic participation in whatever emotional state is depicted onscreen \cite{gallese2019empathic}.  

Popular industry manuals like Arijon's 1976 \textit{Grammar of the Film Language} \cite{arijon1976grammar}  reveal the challenge of translating camera distance---the distance from camera to subject---into computer vision models. ``The gradation of distances can be infinite,'' writes Arijon. ``Actual practice has taught that there are five basic definable distances \ldots However, these denominations do not imply a fixed measurable distance in each case. The terminology is quite elastic, and deals mainly with concepts'' \cite[16]{arijon1976grammar}. This elasticity is largely owed to the fact that the human body serves as the implicit measuring stick across all distance categories, which identify how much of it is visible in the frame. Accordingly, a ``full shot'' of a man and a ``full shot'' of a penguin will each depict the subject’s entire body, but the classification indicates the framing rather than the distance between the camera and the subject, which is different for each. The potential for ambiguity increases when the subject of the shot is unclear, or when it is an inanimate object instead of a human being or another living creature. Additionally, nonstandard lenses like the wide angle and the telephoto exaggerate and compress depth, making it difficult to understand physical camera distance without access to production documentation. 

Accordingly, rather than attempting to assess the (unknowable) physical distance between a camera and its subject, we assess the \textit{apparent} distance---i.e., how far away in feet the subject appears to be. We define ordinal categorical values of extreme close-up, close-up, medium close-up, medium shot, medium long shot, long shot, and extreme long shot, along with a non-ordinal category for intertitles (dialogue text written onscreen, appearing in silent films). Six of those categories are illustrated in figure \ref{fig:distanceexamples}, and our annotation guidelines, which differentiate the boundaries between categories, can be found in \S\ref{taxonomy} (SI).

We annotate a random sample of 3 films per year over the entire period of our study (1922-2025; 312 films) and an additional 100 animated movies, selecting up to one minute from each film to annotate; we segment that clip into shots using TransNetV2\ \cite{soucek2020transnetv2}, and label the exact region of each distance category present in each shot (a total of 4,939 shots).  For shots that involve camera movement (where, for instance, one may begin as a medium shot but transition through a zoom or dolly-in to a medium close-up), we label the start and end point in which a distance category applies. Every frame therefore belongs to exactly one distance category.

Our goal is to scale up an algorithmic measuring instrument to reproduce these human judgments of distance. We train and evaluate several classes of image classification models, sampling 4 frames per second from each shot for both training and evaluation.  The best performing model, base SigLIP2\ \cite{tschannen2025siglip2multilingualvisionlanguage} with $384 \times 384$px resolution, achieves an accuracy of 68.9\% ($\pm 3.3$) and a within-one accuracy of 95.61\% ($\pm 1.0)$; mistakes are generally at the boundary between adjacent distant classes.  Full model comparison can be found in \S\ref{si:modelcomp} (SI).

We use this trained model to predict shot distance over all movies in our collection (again sampling frames at 4 fps).  The figures reported in this article display the raw prevalence of each category. In order to ensure that we accurately reflect prediction error in these prevalence estimates, we also adopt the methodology of prediction-powered inference introduced by Angelopoulous et al.\ \cite{ppi}; in particular, we apply prediction-powered mean estimation and corrected confidence sets for the prevalence of each distance category, using model predictions over our labeled data (where the truth is known) to correct prevalences measured over the unlabeled set.  Because predictive accuracy can vary over time, we correct the prevalences of a category for films in a given year $i$ using corrected predictions over a 25-year window centered on that year $[i \pm 12]$. The resulting corrections can be found in figs \ref{fig:ppimiddle} (for the well-attested middle categories of MCU, M, ML and L) and  \ref{fig:ppiedge} (for rarer categories of XCU, CU, XLS and INTER).  We see the broad trends remaining the same over the entire collection even with this correction, with slightly wider confidence sets reflecting increased uncertainty in prediction error.

\begin{figure*}[t!]
\centering
\includegraphics[width=1\linewidth]{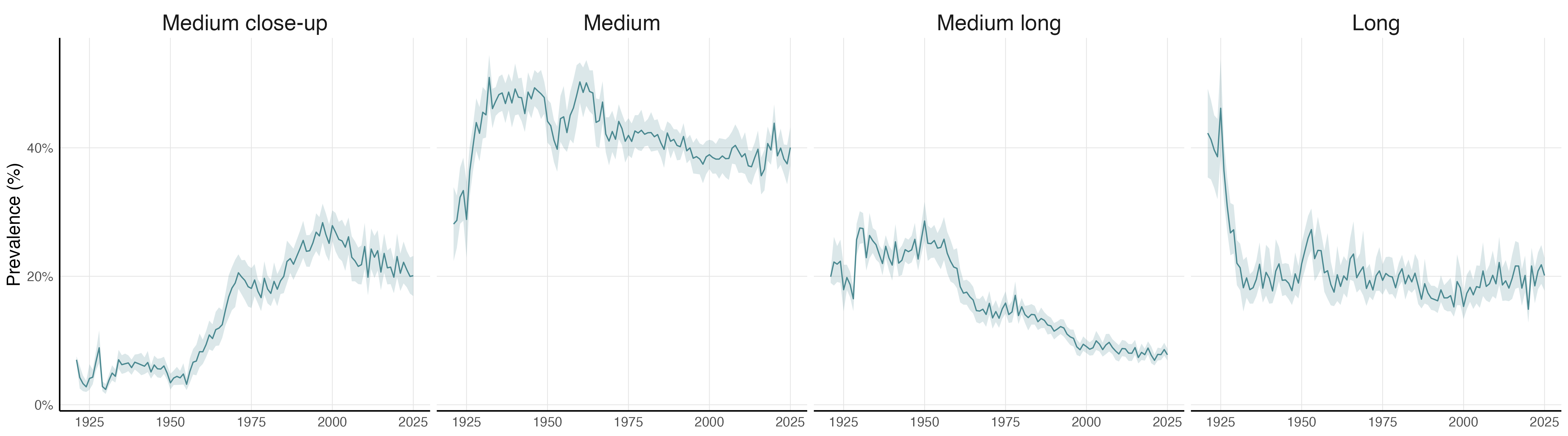}
\caption{Shot distances over time (popular films), along with 95\% confidence intervals (resampling complete movies).}
\label{fig:middistance}
\end{figure*}

As an additional check, we compare the predicted distance proportions for all of the films in our collection that overlap with those that Barry Salt annotated in his \emph{Film Style \& Technology: History \& Analysis}\ \cite{salt2009film}, a total of 89 movies.\footnote{We use published data tables available here: \url{http://www.starword.com/Data_Method/Data_Tables/data_tables.html}.}  We map Salt's distance taxonomy onto our own (BCU $\rightarrow$ CU, CU $\rightarrow$ MCU, MCU $\rightarrow$ M, MS  $\rightarrow$ M, MLS $\rightarrow$ ML, LS $\rightarrow$ L, VLS $\rightarrow$ XLS).  For every film, we calculate the Spearman rank correlation coefficient $\rho$ for all 7 ordinal categories. Across all films, we find the average $\rho = 0.880$, suggesting high agreement between the predicted proportions output from our model and those that Salt created by hand. 

\section{Findings}

Shot distance predictions at this scale allow us to test several hypotheses about the development of apparent camera distance over the past 100 years.  We center these questions around changing temporal dynamics over this period of time, identifying the subjects of the medium close-up in particular, and investigating how animated films employ more flexibility in camera distance due to their constructed nature.

\subsection{Camera distance over time}

Several major technological changes in media production and viewing took place over our period of study from 1922-2025.  In particular:

\begin{itemize}

\item The introduction of sound in the motion picture industry was a fast transition, with nearly all movies moving from complete silent to complete sound over a three-year period between 1928--1931 (see fig \ref{fig:afi}, SI).

\item The US census shows the number of US households with a TV rising from 9.0\% in 1950 to 64.5\% in 1955 to above 97\% for the first time in 1975.\ \cite{us_census_bureau_1980_statistical}; see fig. \ref{fig:tvpenetration} (SI) for full distribution.

    \item Pew reports that the number of smartphone users who used their devices to watch streaming video content (e.g., Netflix, Hulu, etc.) doubled between 2012 and 2015, from 15\% to 33\%\ \cite{anderson2016smartphone}.
\end{itemize}

While we cannot specify a causal relation of the impact of these innovations on movies, we have good reason to think the visual form of film might respond to these changes.  Bordwell et al.\ \cite{bordwell1985classical} identify a temporary decline in camera mobility in the transition period from silent to sound movies resulting from the need to isolate camera noise produced during the simultaneous recording of image and sound (an effect dramatized in the movie \emph{Singing in the Rain}), and Salt, in his manual statistical analysis of ca. 200 films, notes that ``there does tend to be a change in Scale of Shot distribution for most directors between the early sound period, say up to 1932, and the later part of the ‘thirties, as the technical pressures against close shooting are removed''\ \cite[242]{salt2009film}. Conversely, Bordwell notes an increase in close shooting after studios began licensing their libraries to TV stations and selling network broadcast rights for recent releases in the mid-1960s \cite{bordwell2006way}. This led, he argues, to an increase in ``televisionization'' or ``shooting for the box'' as directors anticipated an eventual television release for their films. In the smarphone era, director Paul Schrader offers a similar argument in his assessment of the relationship between filmmaking and technological change, observing that that not only is the increased popularity of the close-up a product of televisionization, but it is also a response to viewing on handheld screens: ``Directors who work in television tell me producers come on set and say: `How are you going to see them on an iPhone? You've gotta come in close'\textquotedblright \cite{schrader2014close}.

Figure \ref{fig:middistance} illustrates the prevalence of the most well-attested shots (medium close-up, medium, medium long and long) in our popular collection over the period 1922-2025 (figures for all shots and all subcorpora can be found in \S\ref{si:prevalence}, SI).  Overall, we see a shortening of camera distance near the introduction of sound, with long shots being replaced by medium shots (and, to a lesser extent, medium long shots). The strongest sustained increase we see is the shortening of camera distance again with the near widespread adoption of television between 1950--1975, where films more than doubled their use of the medium close-up; while the medium shot is largely stable over this period, we see the medium long shot declining.  

We formally test the hypotheses above by identifying the moment when a technological change crosses the 15\% threshold (2012 for smartphone viewing; 1951 for television; and 1929 for sound) and compare the prevalence of each of the well-attested middle distance categories (MCU, M, ML and L) in the three years before that change (e.g., 1926-1928 for sound) compared to 10 years after (e.g., 1936-1938), enough time for the change to lead to reactions in production.  This yields twelve tests; with a Bonferroni correction for those multiple hypotheses, we see a significant change with the increase of the M shot ($+0.054, p \le 0.001$) and decrease in the long shot ($-0.081, p < 0.001$) with the introduction of sound, and an increase in MCU ($+0.044, p \le 0.001$) and decline in ML ($-0.060, p \le 0.001$) during the rise of television.  We see no significant change with the introduction of smartphone viewing.

As figures \ref{fig:popdetail}-\ref{fig:popprestigeindie} (SI) illustrate, this trend cuts across all subcollections; while indie movies may contain slightly closer shots than contemporaneous popular or prestige movies (fig. \ref{fig:popprestigeindie}), the broad temporal patterns are largely identical across all three subcorpora, suggesting that time itself is the most important determinant on this form of visual style.

We carry out two further tests to isolate the effect of time.  First, we consider genre; as table \ref{tab:mcu-by-genre} (SI) notes, the use of the medium close-up varies significantly by genre, appearing most often in sci-fi, thriller and horror movies; since genre also varies by time (we see fewer Film Noir movies now), we plot the prevalence of each major distance category \emph{within} the top five genres over our period of study (Drama, Comedy, Adventure, Romance and Action). As. fig. \ref{fig:genreplot} illustrates, we see the same broad trends hold within each genre as well.

Second, we consider aspect ratio.  As fig. \ref{fig:aspectratios} shows, the use of aspect ratios has changed considerably over the past century of film, with early films largely using a 1.33:1 and 1.37:1 (Academy) ratio,  2.35:1 ratio rising mid-century, 1.85:1 (Flat widescreen) dominating the 1980s--90s, and 2.39:1 commonly used today.  Since these formats may convey different apparent distances, we measure historical prevalence for the major distance categories within each aspect ratio.  As fig. \ref{fig:prevaspectratios} illustrates, we again see the same broad trends hold within each aspect ratio as well.  

\subsection{The male gaze: Women are seen through more medium close-ups than men}

Next, we consider the question: who gets a medium close-up?  While past work has established that men as characters get three times more screentime than women in film\ \cite{bammanpnas}, we have reason to think that medium close-ups (and shorter camera distances in general) will focus on women. Psychoanalytic and screen theory traditions argue that Hollywood cinema is structured around an asymmetry of looking, with men looking and women being looked at. Within this framework the close-up is identified as a key site for the enactment and naturalization of such asymmetry \cite[22]{Mulvey1989}.  Previous computational studies have found such disparate attention given to women in \emph{Bewitched} and \emph{I Dream of Jeannie}\ \cite{arnold2019visual} over their male counterparts; and Hollywood has inherited gendered conventions around the style of close-up from classical portraiture\ \cite{keating2006portrait}.

To test the degree to which we do see women's faces as fragmented objects of attention to a greater degree than men, we recognize characters present on screen using the methodology of past work\ \cite{bammanpnas}, which matches detected faces in a movie to images from its cast list in order to identify the visible actors.  We identify the perceived gender of actors using Wikidata as an approximation to gender knowledge of viewers, temporally scoping an actor's gender to the year of the movie's release (in order to reflect an actor's changing gender perception).  We measure shot distance in frames that contain exactly one recognized actor whose gender is known, and calculate the rate at which men and women appear in different shot distances.

\begin{figure}
\centering
\includegraphics[width=.6\linewidth]{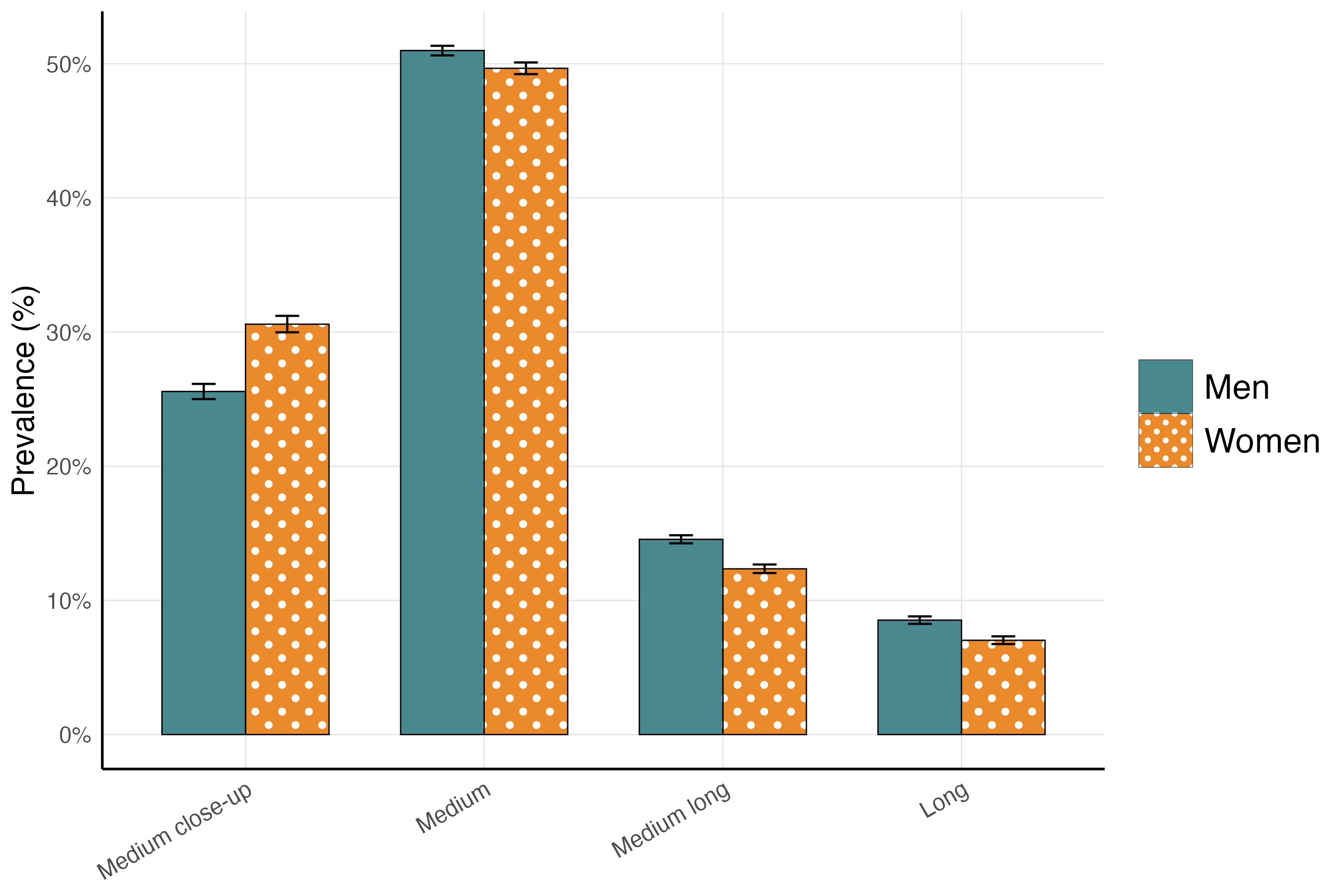}
\caption{Gender rate, popular films; women have 20\% (+5pp) more medium closeups than men ($p < 0.001$ under permutation test).}
\label{fig:gender}
\end{figure}

Figure \ref{fig:gender} and table \ref{tab:gender-rate-popular} (SI) illustrate the result, showing the prevalence of each distance category by gender.  Despite having significantly less screentime than men, women are the subject of medium close-up shots 20\% more frequently (an absolute difference of +5 percentage points; $p < 0.001$ under a two-sided sign-flip permutation test), and show decreasing place in all longer-distance shots as a result.
As figs \ref{fig:genderpoptime}-\ref{fig:genderalltime} (SI) show, we do not see major differences between the different subcollections (popular, prestige or indie); we see slightly decreasing effects over time (with stronger differences in earlier cinema), but the disparity persists throughout the entire period of study.
Close attention to women is a constant throughout this century of film.

\subsection{Animation shows longer distance, and no gendered effects}

Finally, we consider the use of shot distance in animation.  Unlike live-action films, where a physical camera is positioned to capture a subject placed in front of it, animation constructs the entire world that is represented; there is no camera that needs to be constrained by the technological limits placed on live-action production.  In this more liberated medium, do we see camera distance being deployed in the same way as in live-action filmmaking?

Since animated films tend to show up in the top box office lists only later in our collection (80\% of the animated films in our collection were released after the year 2000), we carry out a controlled experiment pairing each animated film with a live-action film from the same genre and year of its release in order to rule out any generic or temporal bias.  We then calculate the distribution of shot distances seen in this paired collection.

\begin{figure*}[htbp]
    \centering
    \includegraphics[width=1\linewidth]{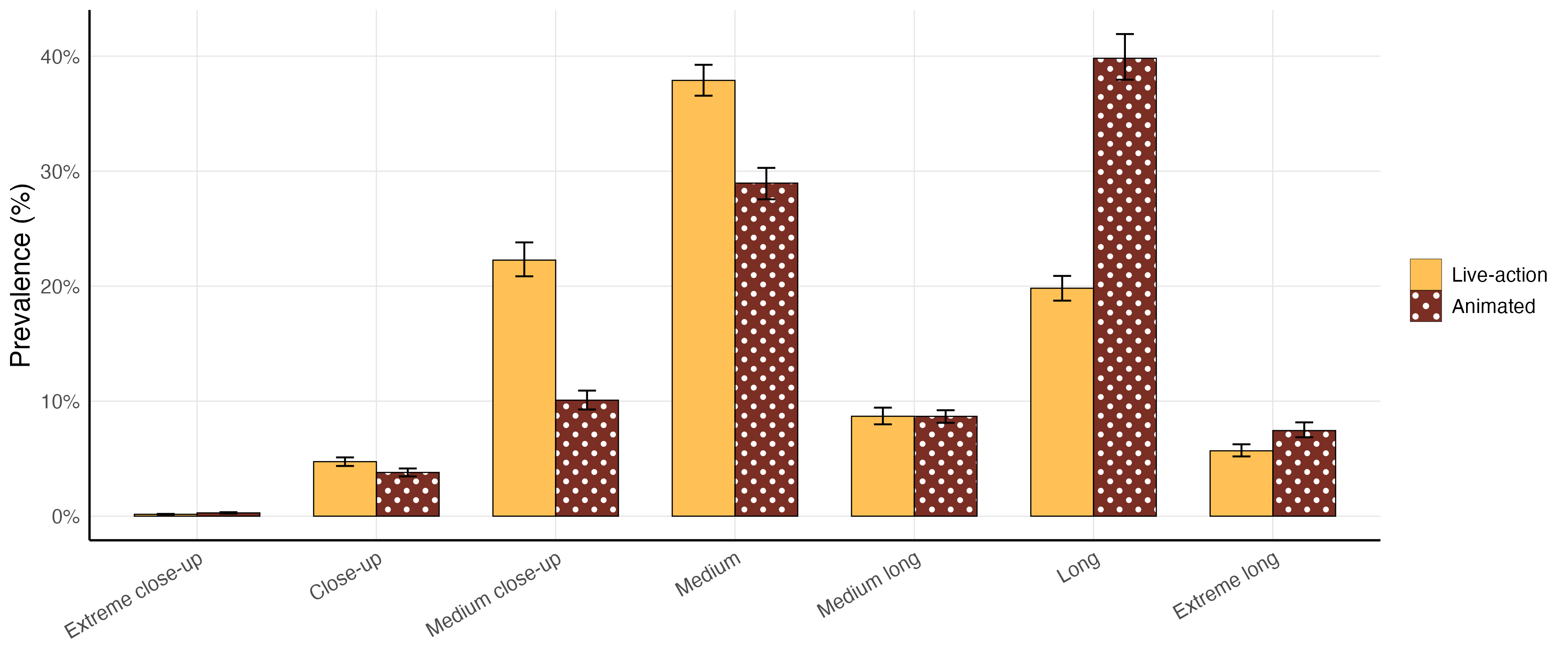}
    \caption{Distribution of shots in paired popular live-action and animated movies.}
    \label{fig:animation}
\end{figure*}

As figure \ref{fig:animation} shows, animation behaves very differently from live-action film.  While live-action films concentrate attention on the medium close-up and medium shot (emphasizing the human face), animation contains over twice as many long shots compared to live action.  While this is partially explained by the freedom from technical limitations in positioning the camera, it also serves to highlight a common subject in animated films: situating a small non-human character (such as a rat in \emph{Ratatouille} or meerkat in \emph{The Lion King}) within their larger environment by drawing attention to the difference in scale.

We additionally measure whether we observe the same difference in gender with respect to the close-up in animated films as we do for live action. While our investigation into gendering of close-ups in live-action films is motivated by past work on the objectification of women on screen, would we expect to see the same in animated films, ostensibly created for a very different audience?

To test this, we recognize the animated characters present on screen using the methodology of past work\ \cite{bamman2026gender}, which trains models for animated character face detection and recognition, matching detected characters against a manually constructed cast list for each film. As in that work, we attribute a character's gender based on the gender of the voice actor (using Wikidata), and draw on fan community discussions to identify characters whose gender is misaligned with the actor (e.g., Bart Simpson, who is voiced by a woman but referred to as a boy).  Since animated characters come in a wide variety of forms (human, animal, machine, etc.), we additionally manually label whether each character has a human or non-human form.

In the paired test, we calculate the difference between female shot prevalence and male shot prevalence for all categories in three conditions: live-action film characters, human animated characters, and non-human animated characters.  As fig. \ref{fig:pairedanimation} (SI) shows, we see a significant difference for medium close-up, medium, and long shots among live-action film characters; we do not see a statistically significant difference in closer shots between male and female characters (for either human or non-human forms) for animation.  The gender difference we see for the medium close-up above is strictly a finding for live-action film.

\section{Discussion}

Our measurements of camera distance over time give us confidence in our methods by confirming two existing accounts in film history. Both of these---the effects of the introduction of sound and of the widespread adoption of television---highlight the intertwined relationship between the representation of onscreen space and technological progress. This is perhaps unremarkable for an art as technologically determined as cinema, yet our third, negative finding of no change following the introduction of the smart phone suggests Schrader's conviction that camera distance conforms to the demands of technology was misplaced. In fact, we see an interesting counterexample of this in a statistically small but nonetheless meaningful subset of mid twentieth-century films that represent the film industry's reaction to television’s widespread introduction into American society. Belton\ \cite{belton1992widescreen} identifies widescreen formats like Cinerama and VistaVision, which date to this period, as a strategy to offer viewers an immersive, spectacular alternative to the small, flat, domestic television screen, and indeed it is films like \emph{This is Cinerama} (1952) that have the greatest proportion of extreme long shots in our data, as widescreen productions favored landscape and travel imagery over close views.

Recent years have seen a similar strategy adopted to lure audiences back to theaters in the post-pandemic era with films like \textit{Oppenheimer} (Christopher Nolan, 2023), \textit{The Odyssey} (Christopher Nolan, 2026), \textit{Sinners} (Ryan Coogler, 2025) and \textit{One Battle After Another} (Paul Thomas Anderson, 2026) embracing IMAX and resurrecting VistaVision \cite{delbarco2025vistavision}. Our finding that smart phones have had no statistically relevant effect on the evolution of camera distance suggests that film grammar may be more resilient across platforms than previous generations of researchers thought. It is also possible that the trend to closer views noted in past work \cite{bordwell2006way,schrader2014close} has played out and can now be considered platform agnostic. Or, as Casetti argues in \cite{casetti2012relocation}, the digital era's effects on cinema are more about changes in the viewing experience and environment than the migration or evolution of formal properties like camera distance. 

Our findings on the gender effects of camera distance---with women receiving far more medium close-ups than men---confirm influential but empirically limited structural claims about the asymmetry of looking. They demonstrate that those claims, made primarily in relation to classical Hollywood cinema, extend beyond a single industrial logic or a particular ideological moment and apply to a significantly longer period extending through to the present day. Our findings also suggest that gender asymmetry in cinema has become normalized to the point that it is essentially the default in live-action films, even in independent cinema, whose purpose is to work against convention. That we see such a strong effect for live-action film but not animation speaks to how this effect is likely rooted, at least in part, in the objectifying function of closer shots. Finally, our findings underscore the value of developing machine learning and computer vision approaches in collaboration with humanists since, in this case at least, analytical AI has served to identify an asymmetry even more systemic and enduring than its original theorists might have imagined, better positioning a new generation of humanists to explore the mechanisms that continue to produce it. 

Taken together, these findings illustrate the affordances of computational research for complementing close viewing of film narrative\ \cite{arnold2023distant}, opening up a range of other narrative phenomena to be measured beyond Salt's early large-scale work on average shot duration\ \cite{d0d389e1-a831-31fa-95b0-d6e44511b8e9}; while this work focuses on shot distance, the same methods naturally extend to other visual forms of film language (such as camera movement, angle, and blocking).  While this work necessarily involves a formal operationalization of these concepts and risks simplifying their inherent complexity (including the creative ways in which filmmakers play with, and resist, standard approaches to the representation of space), it offers a way to test claims about cinematic form that have so far only been addressed at a smaller scale.

\section{Materials and Methods}

\subsection{Data}

For our popular collection, we draw films from the top box office lists published by Bamman et al. 2026\ \cite{bamman2026evaluating}; that work mines historical issues of \emph{Variety} magazine from 1922--1979 and extracts weekly box office information for movies at  theaters in major metropolitan areas of the United States. That automatic extraction is highly accurate (reported $\rho$ = 0.961 between rankings of movies measured from automatic predictions compared to human labels) but is naturally limited to those theaters and cities reported by \emph{Variety}. We source DVDs and Blu-rays from online sources (amazon.com, thriftbooks.com) and local stores (Rasputin Music).  Not all films are available for purchase, and many silent films from the 1920s are lost\ \cite{loc_lost_silent_films_2021}. Figures \ref{fig:popular}-\ref{fig:indie} (SI) illustrates the total number of movies digitized by per year for each of our collections; confidence intervals will generally increase the further back in time we go due to fewer films being available from those eras.  We digitize all DVDs using MakeMKV under US 37 CFR 201.40(b)(4); all movies that are used for annotation are digitized a second time under US 37 CFR 201.40(b)(1), which permits viewing short portions of films for research.  All computing over the entire content of films takes place in the Secure Research Data and Compute Platform at UC Berkeley.

\subsection{Shot classification}
We treat shot distance detection as a supervised classification problem, training models to reproduce human judgments about the apparent distance between the camera and its subject.  We annotate a sample of one minute from a total of 412 films (312 evenly sampled across the entire time period of our collection, plus an additional 100 animated films), segmenting each clip into shots using TransNetV2\ \cite{soucek2020transnetv2}.  Annotation was carried out in two stages. First, annotators labeled all shot distances visible in each shot, generally with two independent annotations per shot; in cases of disagreement, a single adjudicator decided on the final labeling.  In comparing the agreement rate of annotators at the multi-label problem, we find an overall pairwise F1 of 67.5, with nearly all of the disagreements occurring at adjacent distances (i.e., a medium shot mistaken for a medium long shot).  To measure we this, we examine the subset of clips where two annotators assign exactly one label (i.e., no movement) and calculate the within-one accuracy as the fraction of all shots where the distance between two annotator's category labels is $\le 1$, finding a within-one agreement rate of 95.4\%.  The second phase applies only to the 165 clips (3.34\%) that were determined to have multiple distances during phase 1; in this phase, the adjudicator labels the specific time span that each distance category applies to, so that every frame in the final dataset is associated with exactly one distance.

We evenly separate the annotated data into training, development, and test partitions, using the development set to select the best-performing checkpoint for each model family. We evaluate several model architectures: SigLIP 2 (Base and So400m, at 224- and 384-pixel resolutions), ConvNeXt-Large, Swin-B, ResNet-50, and ViT-Base and ViT-Large.  As table \ref{tab:modelsummary} (SI) notes, the best performing model (SigLIP 2 base 384px) achieves an accuracy of 68.9\% and a within-one accuracy of 95.6\%---like human annotations, generally making mistakes only at category boundaries.  

\subsection{Actor recognition}
For live-action movies, we use the same methodology as Bamman et al.\ \cite{bammanpnas} to identify the actors present on screen: we detect faces using YuNet\ \cite{wu2023yunet}, assemble faces into face tracks using the IoU overlap\ \cite{bochinski2017high} and generate a representation of the highest-confidence face in each track using InsightFace\ \cite{guo2019insightface}. We apply the same process to images of the cast list from IMDb for each movie, and match each face representation from the film to the best match among the images of the cast. That work reports an AP@50 of 0.869 for face detection, and an F1 score of 0.852 for person identification.  For animated films, we use the same methodology as Bamman et al.\ \cite{bamman2026gender}, which uses a custom trained RT-DETR\ \cite{lv2024rtdetrv2improvedbaselinebagoffreebies} model for animated face detection and a trained DINOv2\ \cite{oquab2023dinov2} model to generate animated face representations, and otherwise apply the same general pipeline (matching each animated face in the movie to the best face among a constructed cast). That work reports an AP@50 of 0.869 for face detection and a Rank@1 identification accuracy of 0.771.

\subsection{Gender}
For live-action movies, we associate actors with gender information using Wikidata; this captures gender perceptions of the Wikimedia community about the actors present in our films, often sourced to publicly reported documents, and categorizes gender beyond a simple binary (including non-binary, genderfluid etc.).  We use yearly snapshots of Wikidata from 2014--2026 to provide temporal scoping of gender within this period so that the gender we associate with an actor is that perceived closest to the film's release. In this method, we capture gender perceptions of the actor, and not perceptions of the character within the diegesis of the film (which may differ).  For animated films, we associate characters with gender by looking first at the Wikidata gender of the voice actor behind each character, and supplement this information with fan references on \url{fandom.com}, as in Bamman et al. 2026\ \cite{bamman2026gender}.

\section*{Funding}
The research reported in this article was supported by the Humanities and AI Virtual Institute (HAVI), a program of Schmidt Sciences. 

\section*{Acknowledgments}

This work was made possible by the use of the Secure Research Data and Compute
Platform at the University of California, Berkeley.

\section*{Competing interests} The authors declare that they have no competing interests.

\section*{Author contributions} D.B., A.C and D.H designed and performed research; D.B. and M.M. annotated data; D.B. and D.H conducted the experiments and analyzed data; D.B. and A.C wrote the manuscript; D.B., A.C., D.H and M.M. reviewed and edited the manuscript.

\section*{Data availability} Some study data are available: we are not able to directly republish the original movies extracted from DVDs or Blu-rays under the current exemption to the DMCA, but we make available several other forms of data to encourage openness and reproducibility. We release our computational pipeline (including trained models for shot distance classification)    so that others are able to run our methods on their own collections; and we release all derived measured that we have calculated in our collection (including the predicted probabilities for all shot distances measured in all films). Code and data to support this work can be found on GitHub (\url{https://github.com/bamman-group/camera-distance}). 

\beginsupplement
\section*{Supplementary Information}

\section{Distance taxonomy}\label{taxonomy}
The following contains the distance taxonomy and reference guide that we use in manually annotating clips:
\\[10pt]
\noindent
When measuring distance, we are largely assessing the apparent distance between the camera and its subject -- i.e., how far away (e.g., in feet) the subject appears to be. We divide this into eight categories (7 ordinal):

\begin{itemize}

\item Extreme close-up: A shot generally taken from under a foot away from the subject. It can depict a detail of a human face (not the face in its entirety), or any other object from a distance of under a foot (e.g., a complete deck of cards).

\item Close-up: A shot generally taken from roughly 1-3 feet away from the subject. For a human subject, this can fill the frame with an entire face. If a face is the subject but does not fill the frame (e.g., if the entire face and shoulders can be seen), then the shot is likely a medium close-up.

\item Medium close-up: A shot generally taken from roughly 3-5 feet away from the subject, often corresponding to a shot of a face (or faces) from the shoulders up. The dividing line between a medium close-up and a medium shot is roughly the chest.

\item Medium: A shot generally taken from roughly 5-7 feet away from the subject, often corresponding to a shot of a person (or people) from the waist up. The dividing line between a medium shot and a medium long shot is roughly the thigh.

\item Medium long: A shot generally taken from roughly 7-10 feet away from the subject, often corresponding to a shot of a person (or people) from the knee up.

\item Long: A shot generally taken from roughly 10-25 feet away from the subject, often corresponding to a shot of the entire body of a person (or people), along with some of the surrounding environment.

\item Extreme long shot: A shot generally taken over 25 feet away from the subject (at least twice the height of a person), often depicting much of the surrounding environment.

\item Intertitles: This category should be used exclusively for text intertitles in silent films.

\end{itemize}

For worlds that are filmed outside of typical human scale -- e.g., shrunken worlds as in \emph{Fantastic Voyage} and \emph{Honey I Shrunk the Kids}, or animated films such as \emph{Rio} and \emph{A Bug's Life} (where the story unfolds at the scale of a bird and bug, respectively), we judge apparent distance with respect to the scale depicted. For example, a shot that depicts a miniaturized character in \emph{Honey I Shrunk the Kids} from head to toe would likely be a long shot, not an extreme close up.

\paragraph{Camera movement/multiple labels.}

For clips involving camera movement, it is possible that multiple distance categories apply (e.g., a medium shot moving to a close-up). In that case, select all distance categories that are held for a sustained period of time (approximately one second or more).

\paragraph{Multiple shots (shot segmentation error).}

If a clip has multiple shots within it, go ahead and label all the distances you see.

\paragraph{Judging scale.}

When we are judging camera distance, we are doing so with respect to an implicit scale---is the scale of what's depicted to be judged from the perspective of a human (viewer or subject) or from (e.g.) a bee (in \emph{Bee Movie})? Or from a miniaturized human scale (as in \emph{Honey I Shrunk the Kids})? Once we establish the scale, we judge apparent distance with respect to it.

Most movies have one scale (for live-action movies this is generally human scale---but not always, as in the case of \emph{Godzilla}); animated movies sometimes jump around between scales when characters are of very different sizes. When judging scale, use your implicit judgment about the dominant subject to guide you.

\FloatBarrier

\section{Total features over time}
\begin{figure}[H]
    \centering
    \includegraphics[width=.9\linewidth]{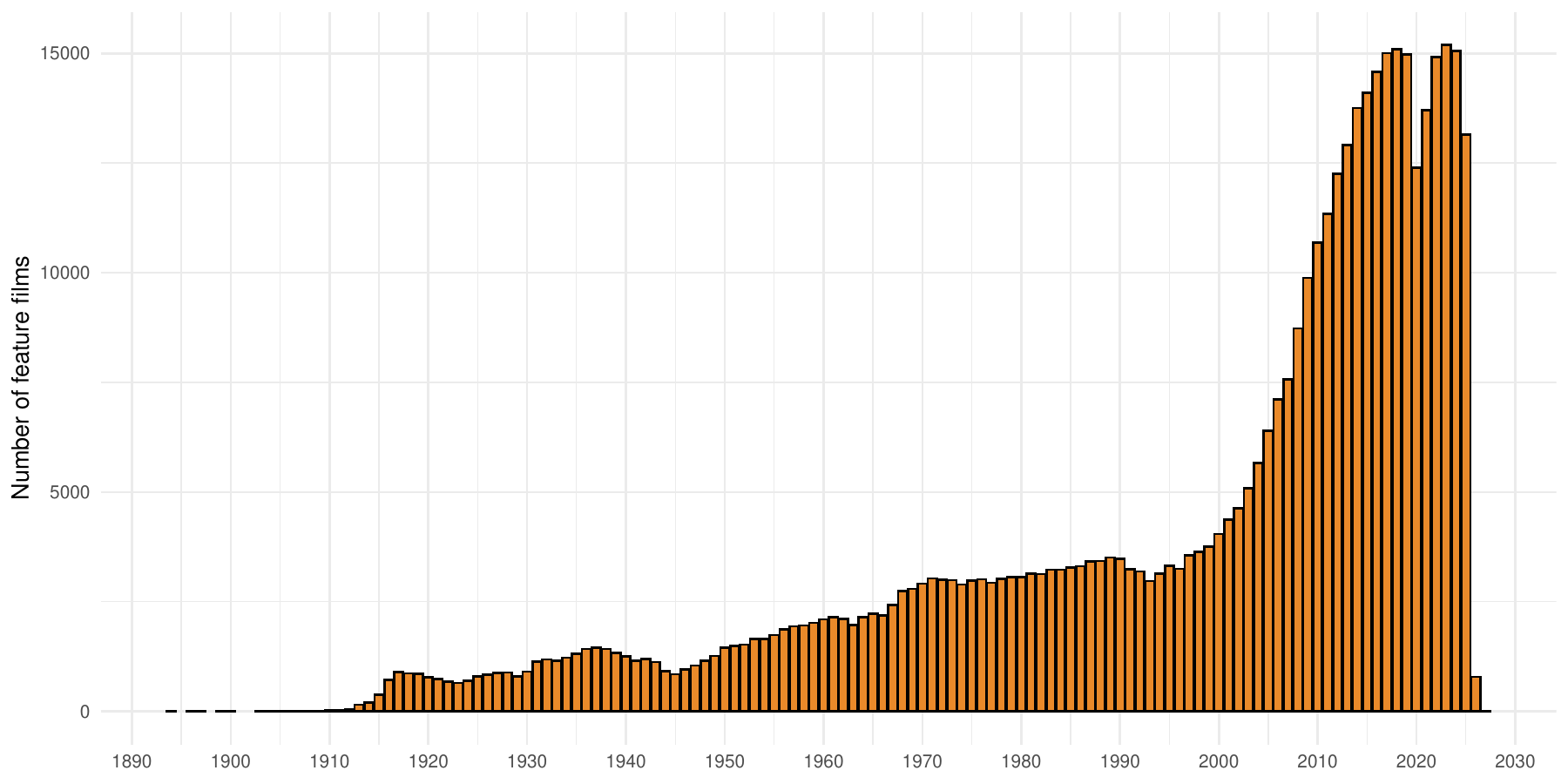}
    \caption{Total number of feature films over time. Source: IMDb.}
    \label{fig:imdbfeatures}
\end{figure}

Figure \ref{fig:imdbfeatures} lists the total number of feature films longer than 40 minutes recorded by IMDb. A roughly linear increase in the number of films produced through the year 2000 turns into an exponential growth in the new millennium.

\FloatBarrier

\section{Size of digitized collection}

\begin{figure}[h!]
    \centering
    \includegraphics[width=.9\linewidth]{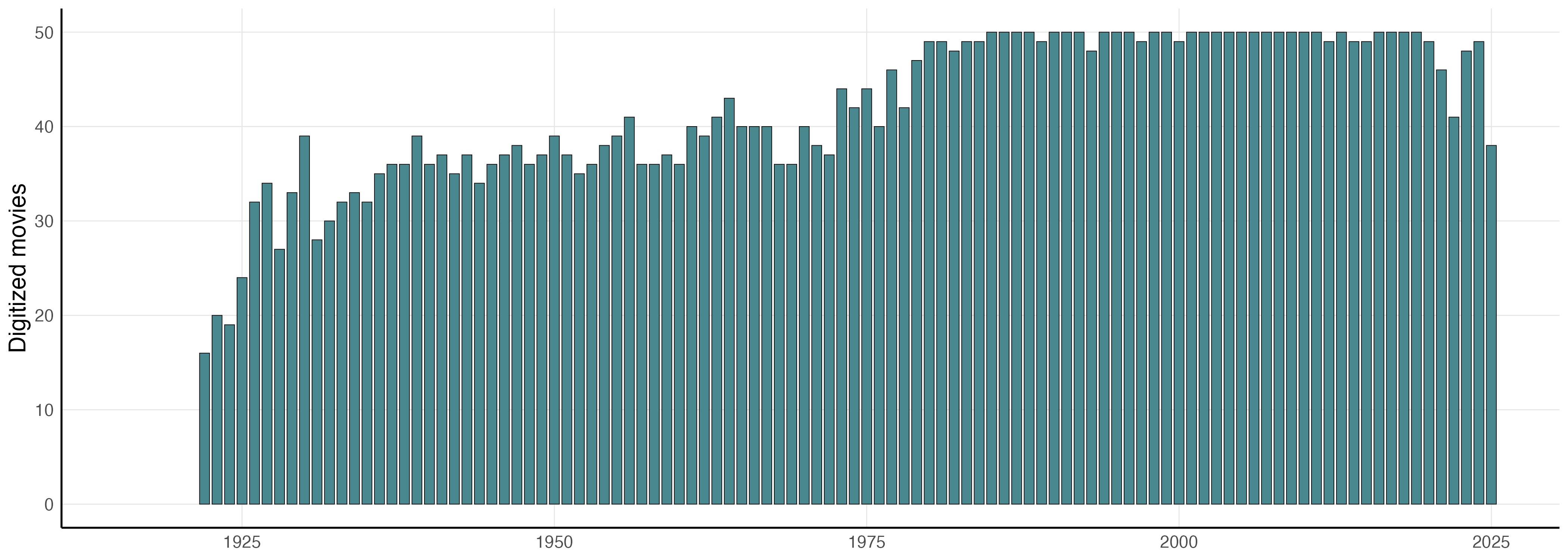}
    \caption{Total number of digitized movies in our popular collection.}
    \label{fig:popular}
\end{figure}

\begin{figure}[ht!]
    \centering
    \includegraphics[width=.9\linewidth]{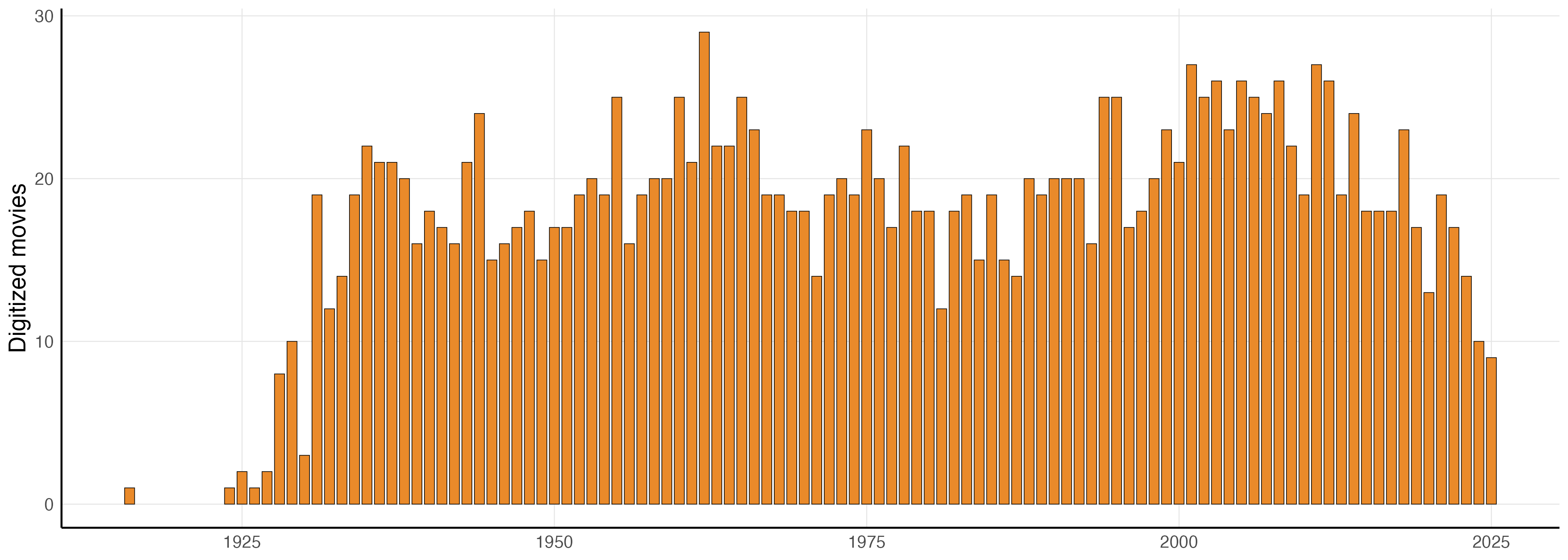}
    \caption{Total number of digitized movies in our prestige collection.}
    \label{fig:prestige}
\end{figure}

\begin{figure}[ht!]
    \centering
    \includegraphics[width=.9\linewidth]{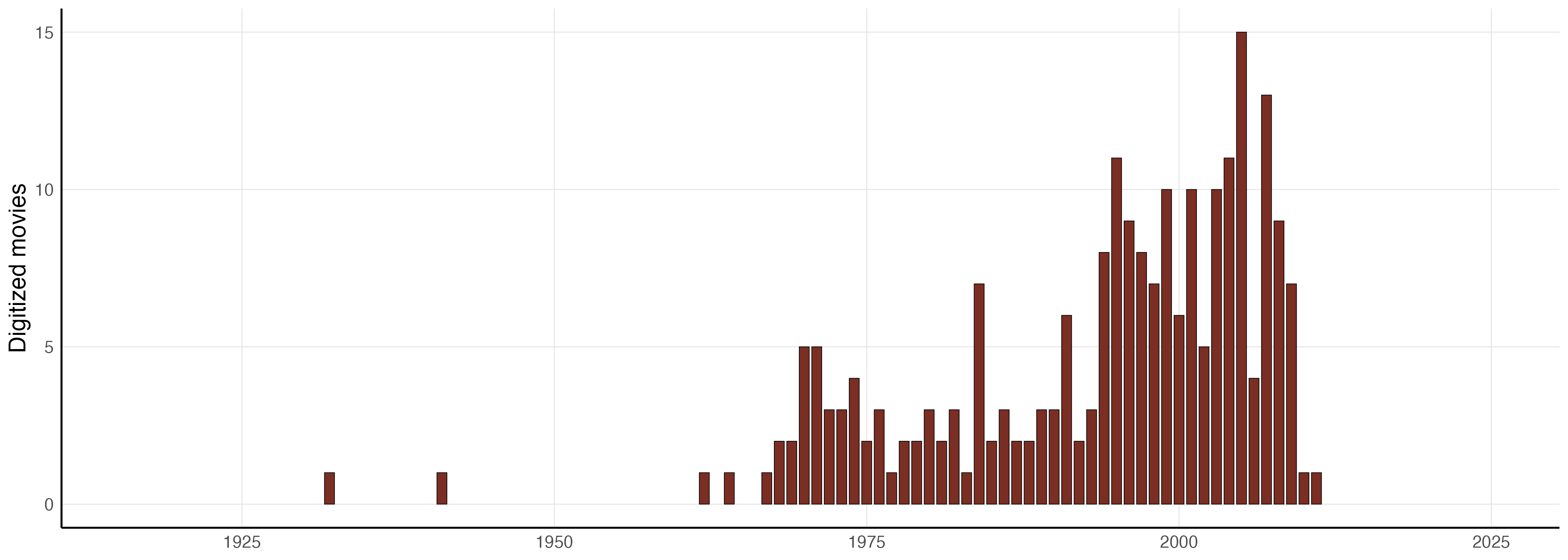}
    \caption{Total number of digitized movies in our American independent collection.}
    \label{fig:indie}
\end{figure}

Figures \ref{fig:popular}-\ref{fig:indie} illustrate the total number of movies per year within our popular, prestige and American independent collections. The collection largely spans 1922-2025, with a single film (\emph{Intolerance}) from 1916 (appearing in AFI and BFI 100).

\FloatBarrier

\section{Technological change}

\begin{figure}[htbp]
    \centering
    \includegraphics[width=.7\linewidth]{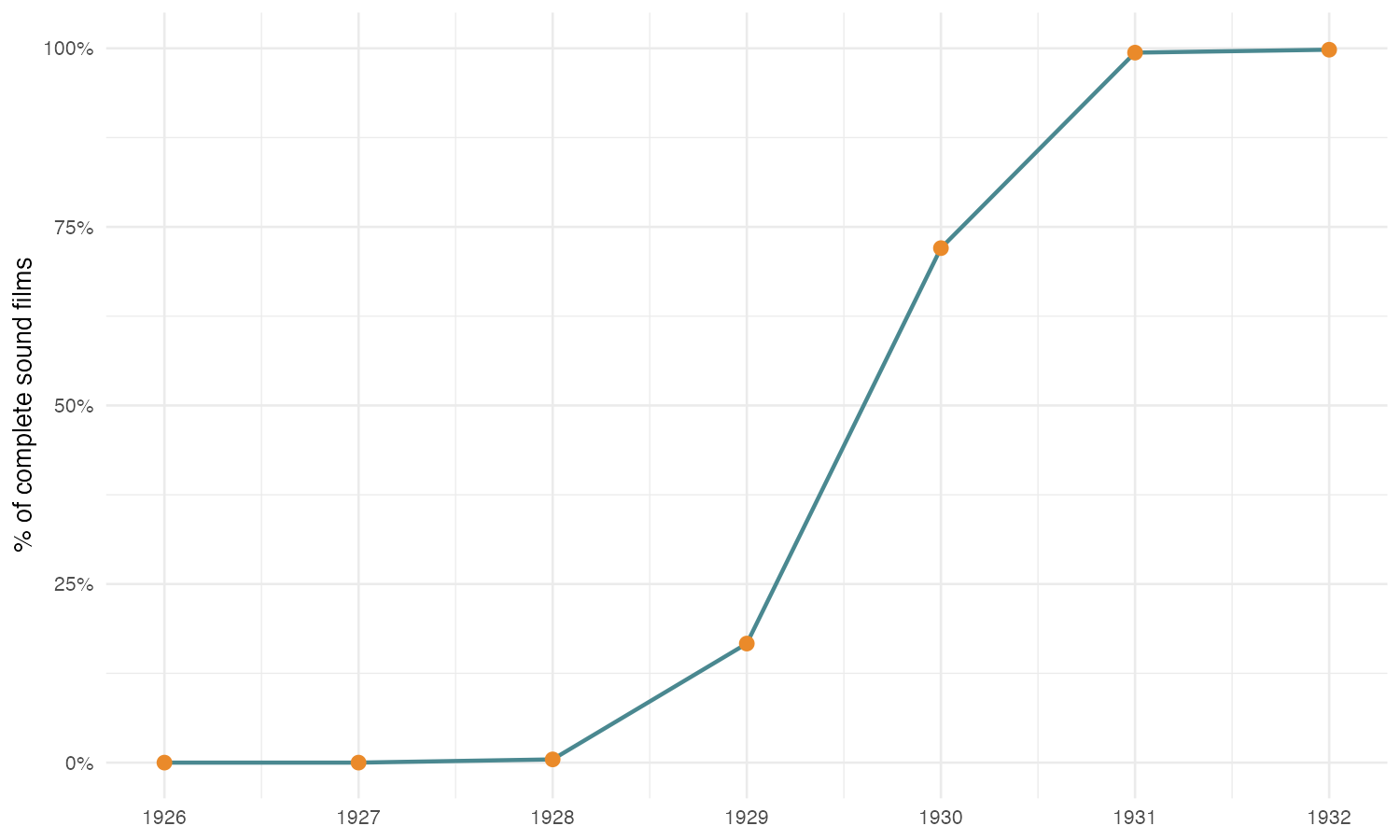}
    \caption{Fraction of movies that are sound, AFI.}
    \label{fig:afi}
\end{figure}

Figure \ref{fig:afi} illustrates the number of movies recorded by the American Film Institute catalog with a physical property of a Sound film among all films with a ``Sound'', ``Silent'' or mixed Sound/Silent (e.g. ``Silent with sound sequences'') property and duration of longer than 40 minutes. With the introduction of the first sound film in 1927, we see a fast transition from nearly complete silent films in 1927 to nearly complete sound films by 1931.

\begin{figure}[htbp]
    \centering
    \includegraphics[width=.7\linewidth]{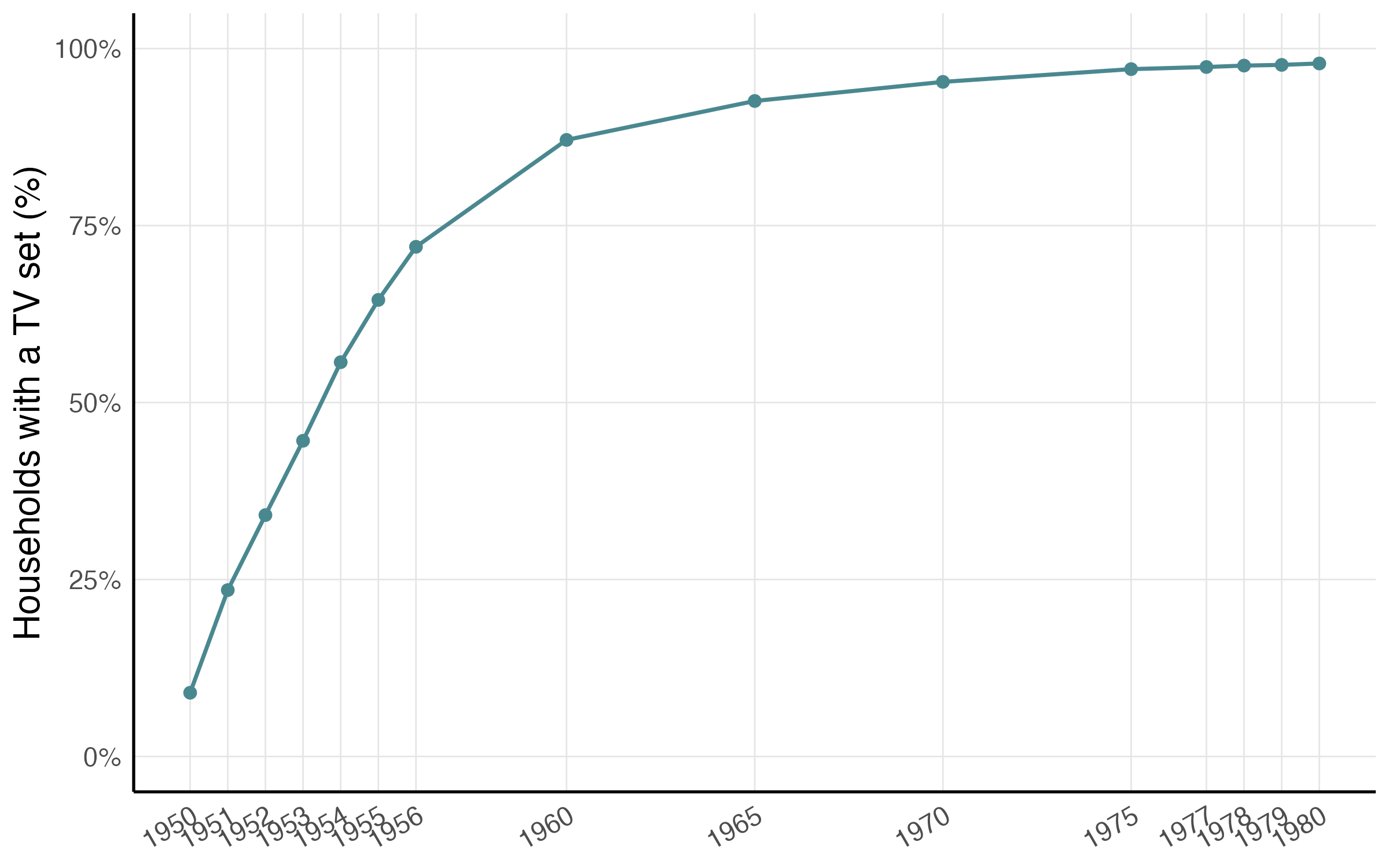}
    \caption{TV penetration in U.S. households. Plotted from data in \emph{Statistical Abstract of the United States: 1980}, table 997 (Households With Television Sets: 1950 to 1980)\ \citep{us_census_bureau_1980_statistical} and \emph{Historical Statistics of the United States, Colonial Times to 1957}\ \citep{us_census_1960_historical}.}
    \label{fig:tvpenetration}
\end{figure}

\noindent
Likewise, figure \ref{fig:tvpenetration} plots the penetration of television into U.S. households over the period 1950-1980, using data from the U.S. census \emph{Statistical Abstract of the United States}\ \citep{us_census_bureau_1980_statistical} and \emph{Historical Statistics of the United States, Colonial Times to 1957}\ \citep{us_census_1960_historical}.

\FloatBarrier
\section{Model comparison}\label{si:modelcomp}

Table \ref{tab:modelsummary} lists the comparative performance across all models we train and evaluate, along with 95\% bootstrap confidence intervals (resampling at the level of complete movies).

\begin{table}[htbp]
\centering
\caption{Model comparison summary}
\label{tab:modelsummary}
\small
\begin{tabular}{lccc}
\toprule
model & accuracy [95\% CI] & within-one accuracy [95\% CI] & training time (min) \\
\midrule
SigLIP 2 (base 224px) & 0.6680 [0.6332, 0.7038] & 0.9497 [0.9381, 0.9619] & 25.3 \\
SigLIP 2 (base 384px) & \textbf{0.6893} [0.6599, 0.7223] & \textbf{0.9561} [0.9461, 0.9661] & 102.4 \\
SigLIP 2 (So400m 384px) & 0.6807 [0.6476, 0.7108] & 0.9382 [0.9237, 0.9514] & 447.9 \\
ConvNeXt-Large & 0.6400 [0.6050, 0.6750] & 0.9358 [0.9193, 0.9502] & 62.5 \\
Swin-B & 0.6516 [0.6161, 0.6857] & 0.9420 [0.9269, 0.9548] & 68.7 \\
ResNet-50 & 0.5984 [0.5629, 0.6339] & 0.9149 [0.8962, 0.9325] & 17.3 \\
ViT-Base & 0.5743 [0.5359, 0.6116] & 0.9161 [0.8986, 0.9329] & 51.1 \\
ViT-Large& 0.6319 [0.5976, 0.6677] & 0.9435 [0.9274, 0.9580] & 92.2 \\
\bottomrule
\end{tabular}
\end{table}

\section{Genre}

Table \ref{tab:mcu-by-genre} lists the prevalence of medium close-ups by genre, using genre information from IMDb (for all genres with at least 30 movies), along with 95\% bootstrap confidence intervals (resampling at the level of complete movies).

\begin{table}[h!]
\centering
\begin{tabular}{lcr}
\toprule
Genre & Medium close-up (\%) & Movies \\
\midrule
Sci-Fi & 25.3 (24.1, 26.5) & 292 \\
Thriller & 24.9 (23.9, 26.0) & 512 \\
Horror & 23.4 (22.1, 24.6) & 287 \\
Action & 22.9 (22.2, 23.6) & 1003 \\
Mystery & 22.0 (20.8, 23.2) & 347 \\
Documentary & 21.2 (19.0, 23.6) & 121 \\
Crime & 20.9 (20.1, 21.8) & 761 \\
Biography & 19.7 (18.5, 20.9) & 372 \\
Fantasy & 19.0 (17.9, 20.0) & 335 \\
Sport & 18.8 (16.6, 20.9) & 97 \\
Drama & 17.1 (16.6, 17.5) & 2997 \\
Adventure & 16.9 (16.3, 17.6) & 1098 \\
Music & 16.7 (14.9, 18.3) & 201 \\
History & 16.0 (14.5, 17.5) & 260 \\
Comedy & 14.1 (13.6, 14.5) & 1897 \\
Family & 13.5 (12.5, 14.5) & 334 \\
Romance & 12.8 (12.2, 13.5) & 1243 \\
War & 12.7 (11.5, 14.0) & 264 \\
Western & 10.7 (8.7, 12.7) & 117 \\
Animation & 10.2 (9.6, 10.9) & 249 \\
Film-Noir & 7.7 (6.6, 8.9) & 70 \\
Musical & 6.3 (5.4, 7.2) & 266 \\
\bottomrule
\end{tabular}
\caption{Share of screen time at medium close-up by genre, with 95\% bootstrap confidence intervals.}
\label{tab:mcu-by-genre}
\end{table}

\begin{figure}[h!]
    \centering
    \includegraphics[width=1\linewidth]{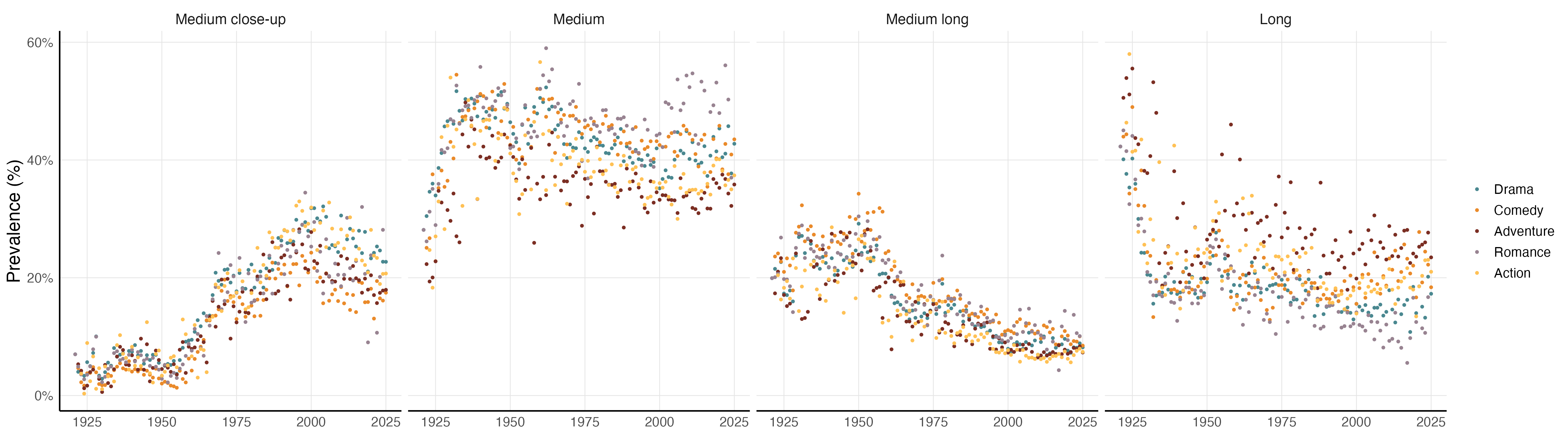}
    \caption{Prevalence (popular) within genres (Drama, Comedy, Adventure, Romance, Action).}
    \label{fig:genreplot}
\end{figure}

\FloatBarrier
Figure \ref{fig:genreplot} illustrates the prevalence of each of the four major distance categories by genre. Plotted as one dot per year per genre for simpler viewing; of note is that no genre displays a strongly different pattern from the overall prevalence.

\begin{figure}[h!]
    \centering
    \includegraphics[width=1\linewidth]{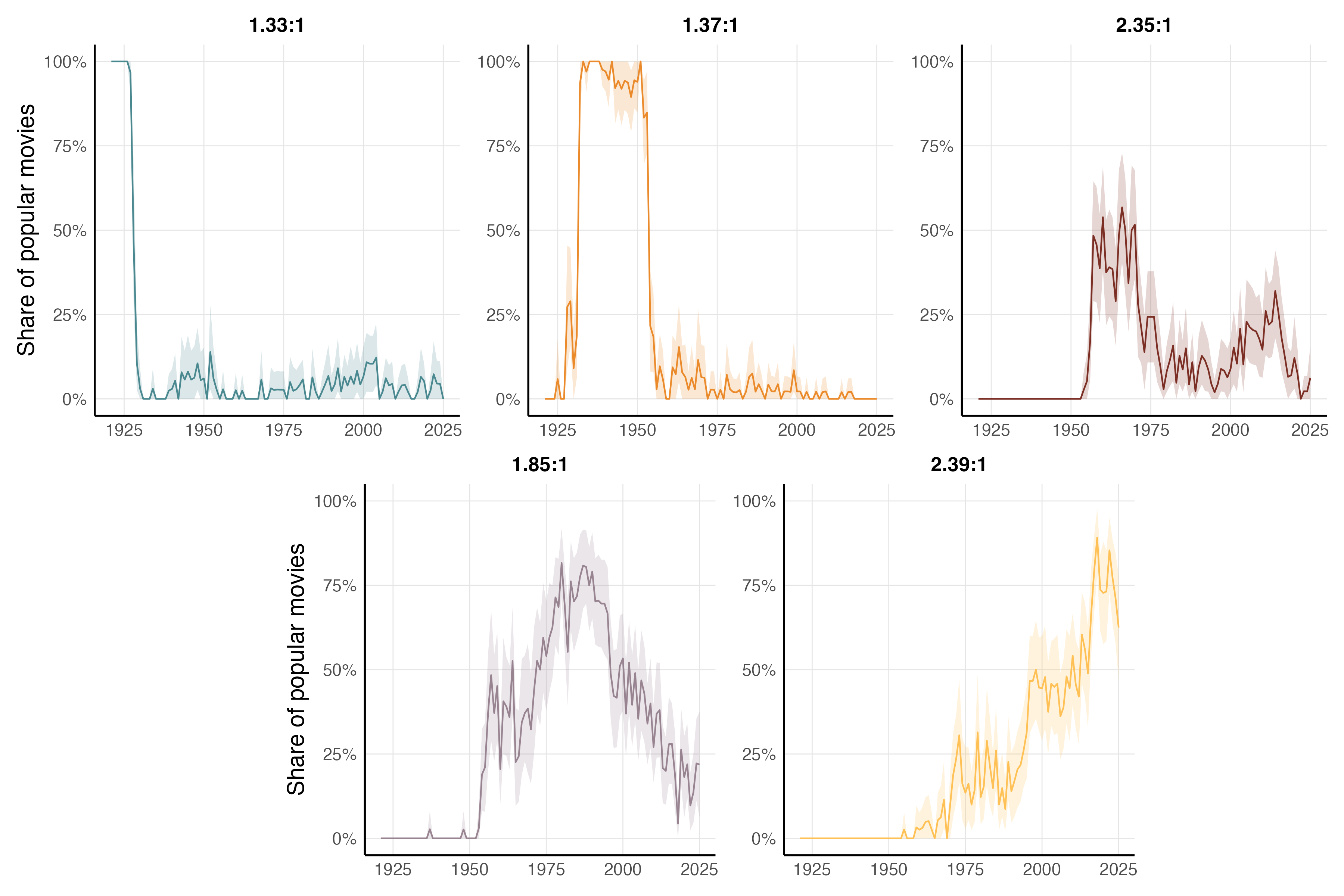}
    \caption{Top 5 most common aspect ratios used in popular films.}
    \label{fig:aspectratios}
\end{figure}

Figure \ref{fig:aspectratios} documents the five most common aspect ratios in our popular film collection; aspect ratio data from IMDb. If a movie is released under multiple aspect ratios, it counts for each one.

\begin{figure}[ht!]
    \centering
    \includegraphics[width=1\linewidth]{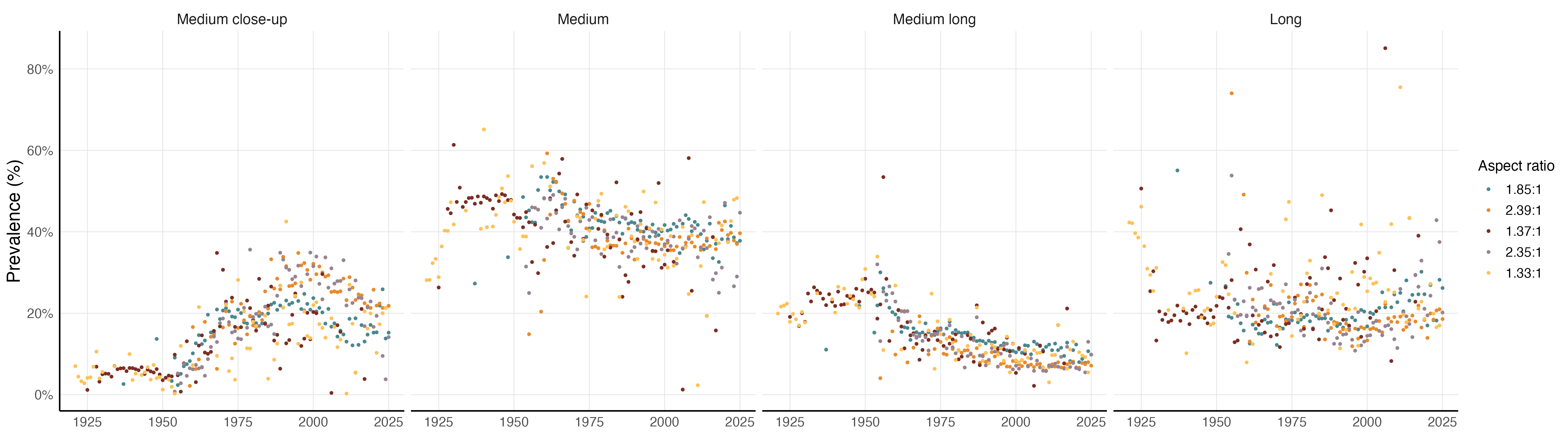}
    \caption{Camera distance over time within each aspect ratio.}
    \label{fig:prevaspectratios}
\end{figure}

\FloatBarrier
Figure \ref{fig:prevaspectratios} illustrates the prevalence of each of the four major distance categories by aspect ratio. Again plotted as one dot per year per aspect ratio for simpler viewing. As figure \ref{fig:aspectratios} notes, aspect ratios are generally dominant at a single period of time, and none persists throughout the entire duration of our collection. Of note is that no aspect ratio displays a strongly different pattern from the overall prevalence; we see a rise in MCU over 1950-1975 and decline in ML over that same period, along with a relatively stable rates of long shots after an initial decline ca. 1930.

\FloatBarrier
\section{Prevalence} \label{si:prevalence}
\begin{figure}[h!]
    \centering
    \includegraphics[width=.8\linewidth]{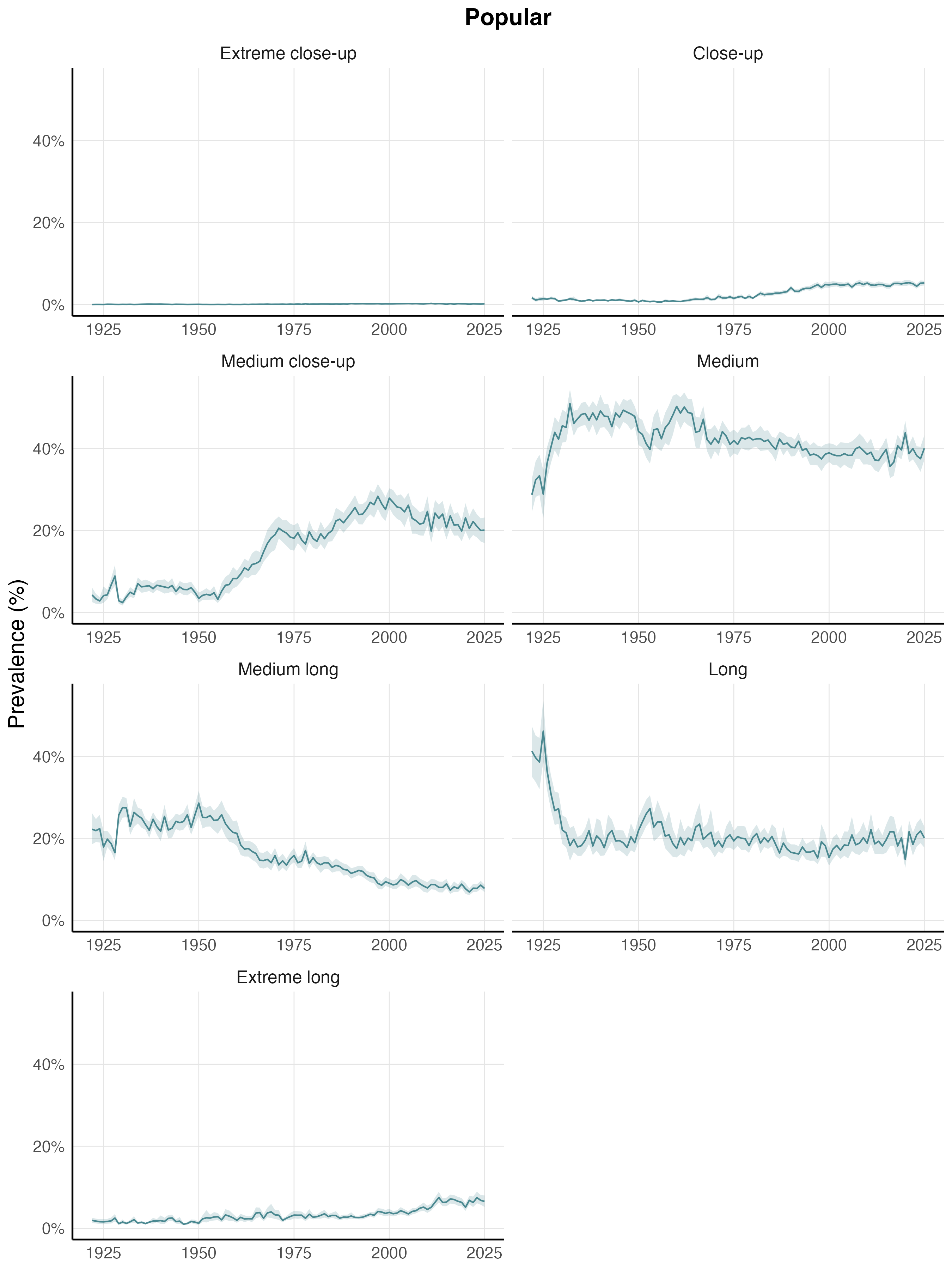}
    \caption{Prevalence of distance categories over time, popular movies.}
    \label{fig:popall}
\end{figure}

\begin{figure}[h!]
    \centering
    \includegraphics[width=.8\linewidth]{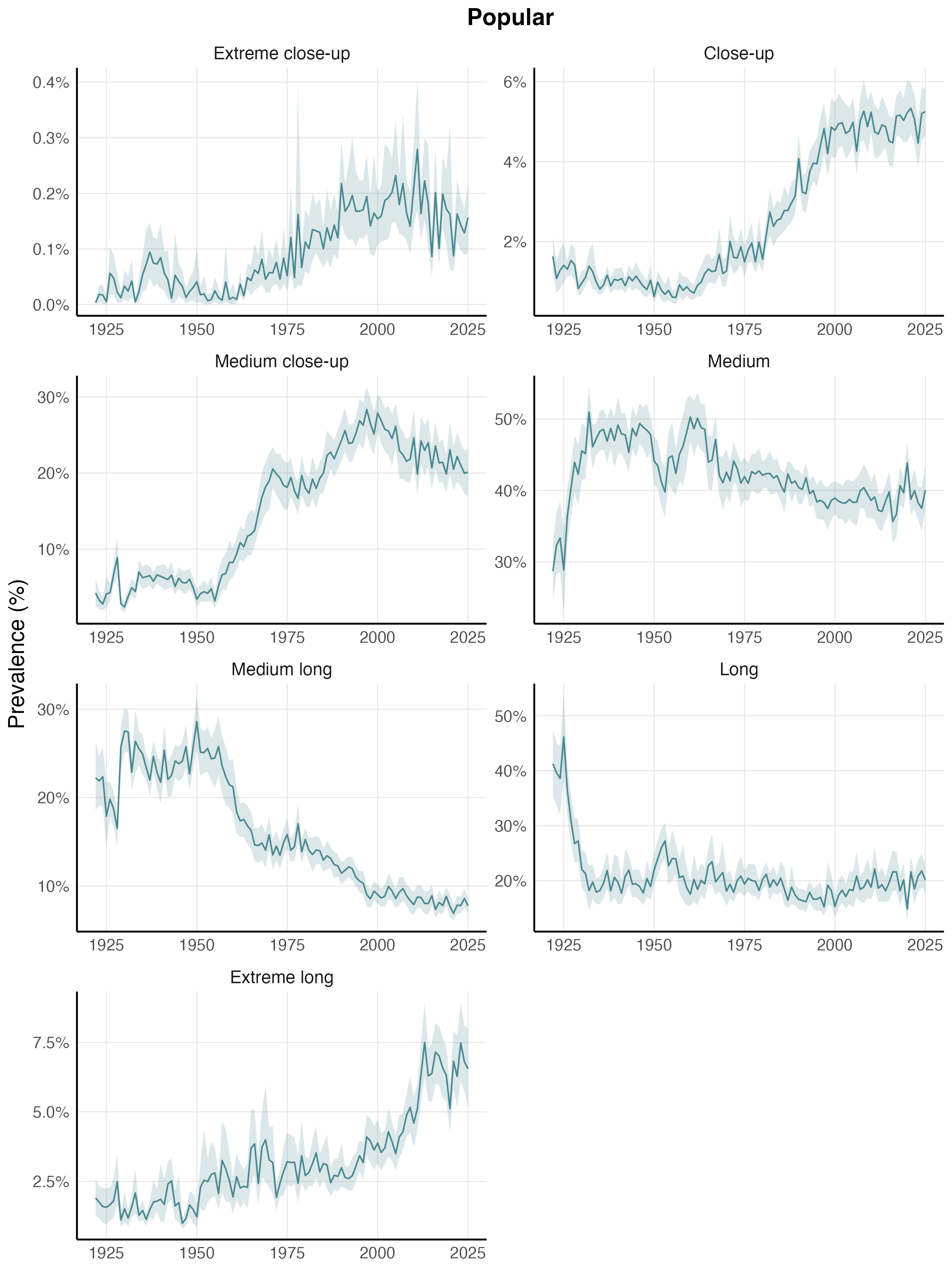}
    \caption{Prevalence of distance categories over time, popular movies (detail).}
    \label{fig:popdetail}
\end{figure}

\begin{figure}[h!]
    \centering
    \includegraphics[width=.8\linewidth]{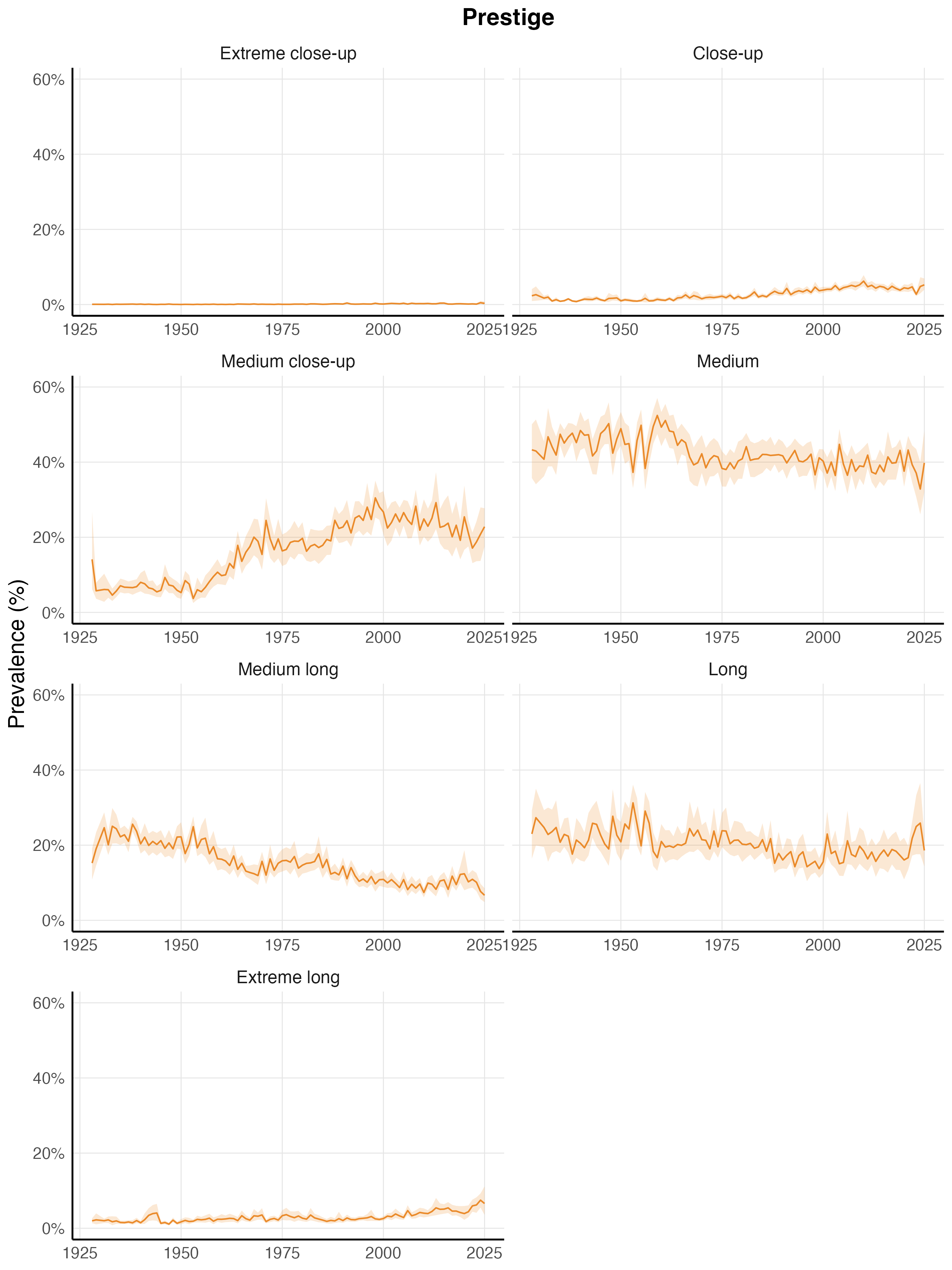}
    \caption{Prevalence of distance categories over time, prestige movies.}
    \label{fig:prestigeall}
\end{figure}

\begin{figure}[h!]
    \centering
    \includegraphics[width=.8\linewidth]{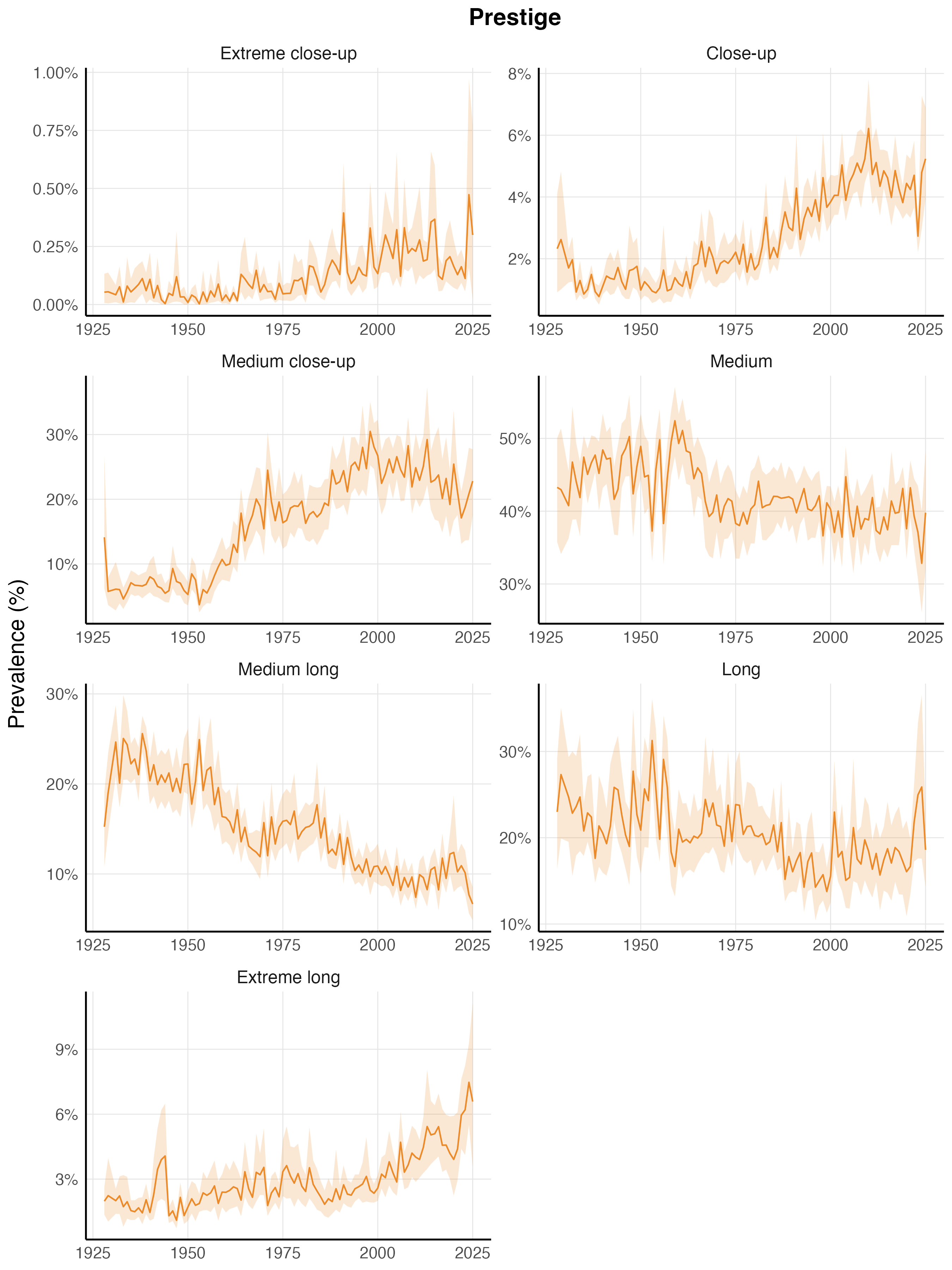}
    \caption{Prevalence of distance categories over time, prestige movies (detail).}
    \label{fig:prestigedetail}
\end{figure}

Figures \ref{fig:popall}-\ref{fig:prestigedetail} illustrate the prevalence of each distance category over historical time, both at a common scale (\ref{fig:popall}, \ref{fig:prestigeall}), and in detail (\ref{fig:popdetail}, \ref{fig:prestigedetail}). All figures show 95\% bootstrap confidence intervals, resampling at the level of a complete movie.

\FloatBarrier
\section{PPI correction} 
\begin{figure}[htbp]
    \centering
    \includegraphics[width=.8\linewidth]{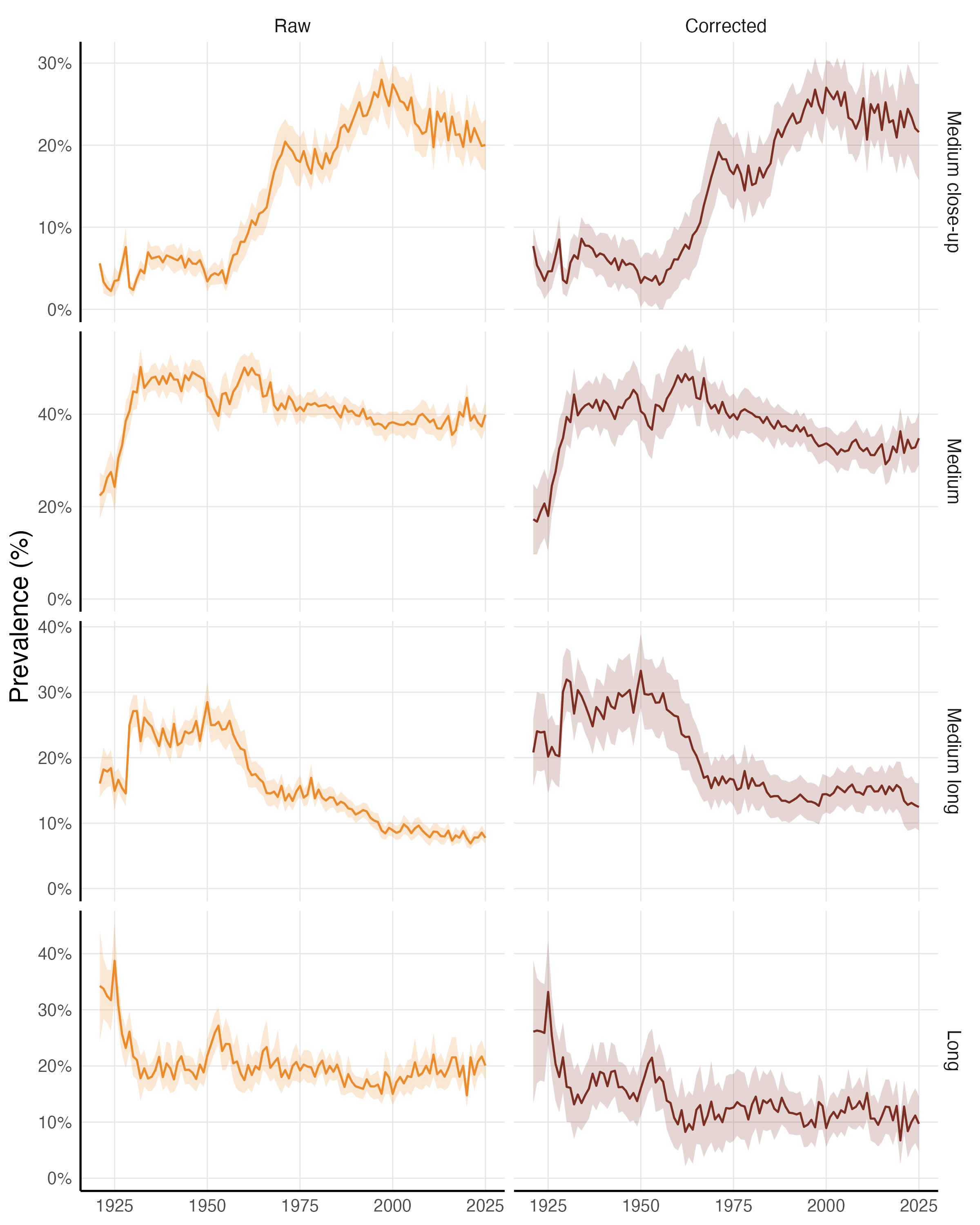}
    \caption{Raw vs. PPI corrected (middle categories).}
    \label{fig:ppimiddle}
\end{figure}

\begin{figure}[htbp]
    \centering
    \includegraphics[width=.8\linewidth]{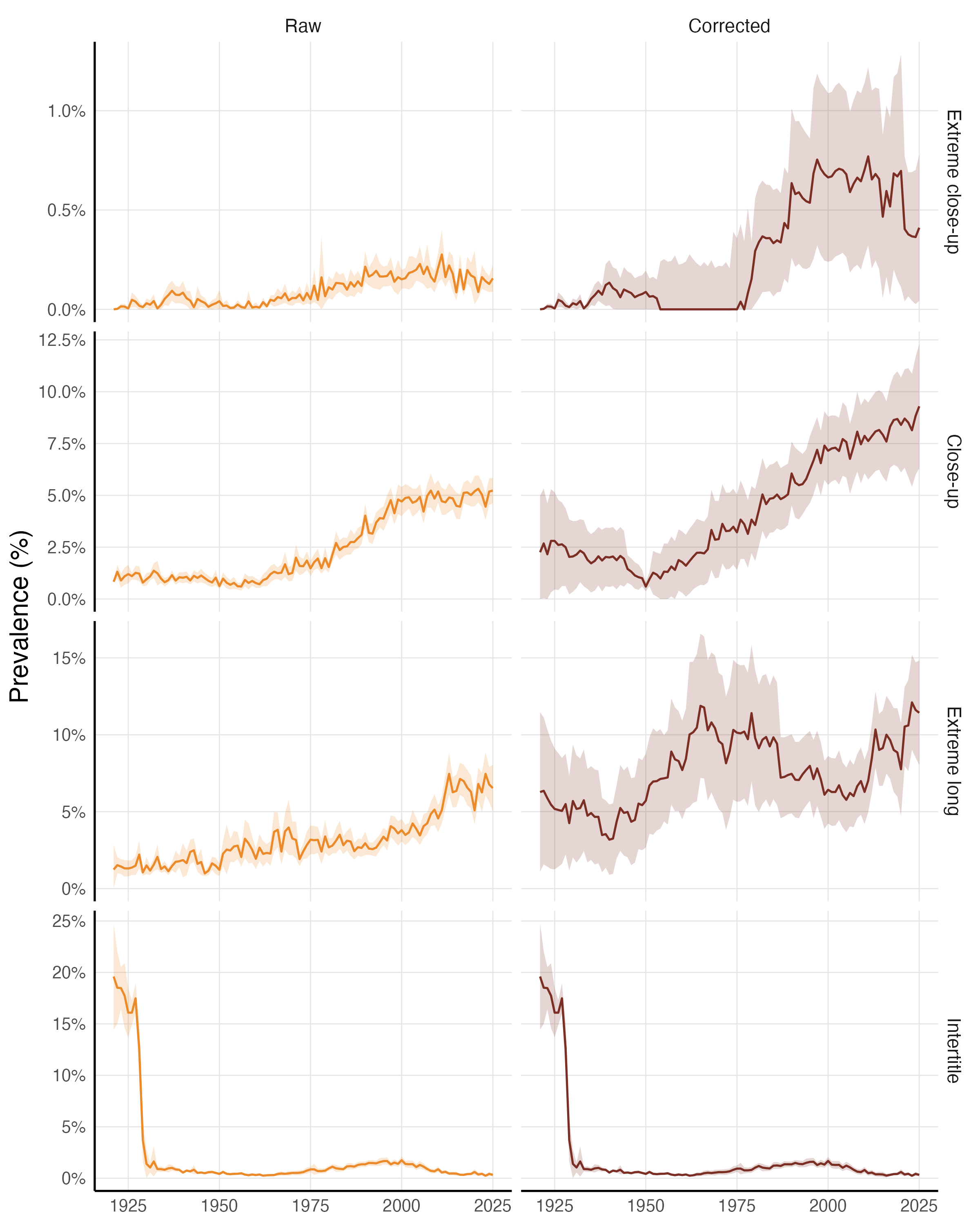}
    \caption{Raw vs. PPI corrected (edge categories).}
    \label{fig:ppiedge}
\end{figure}

In order to ensure that we propagate training error into our final prevalence estimates, we adopt the methodology of prediction-powered inference\ \citep{ppi}. We apply prediction-powered mean estimation and corrected confidence sets over for the prevalence of each distance category, using model predictions over our labeled data (where the truth is known) to correct prevalences measured over the unlabeled set. Because predictive accuracy can vary over time, we correct the prevalences of a category for films in a given year $i$ using corrected predictions over a 25-year window centered on that year $[i \pm 12]$. The resulting corrections can be found in figs \ref{fig:ppimiddle} (for the well-attested middle categories of MCU, M, ML and L) and \ref{fig:ppiedge} (for rarer categories of XCU, CU, XLS and INTER). We see the broad trends remaining the same over the entire collection even with this correction, with slightly wider confidence sets reflecting increased uncertainty in prediction error.

\FloatBarrier
\section{Comparison by subcollection} 
\begin{figure}[htbp]
    \centering
    \includegraphics[width=.85\linewidth]{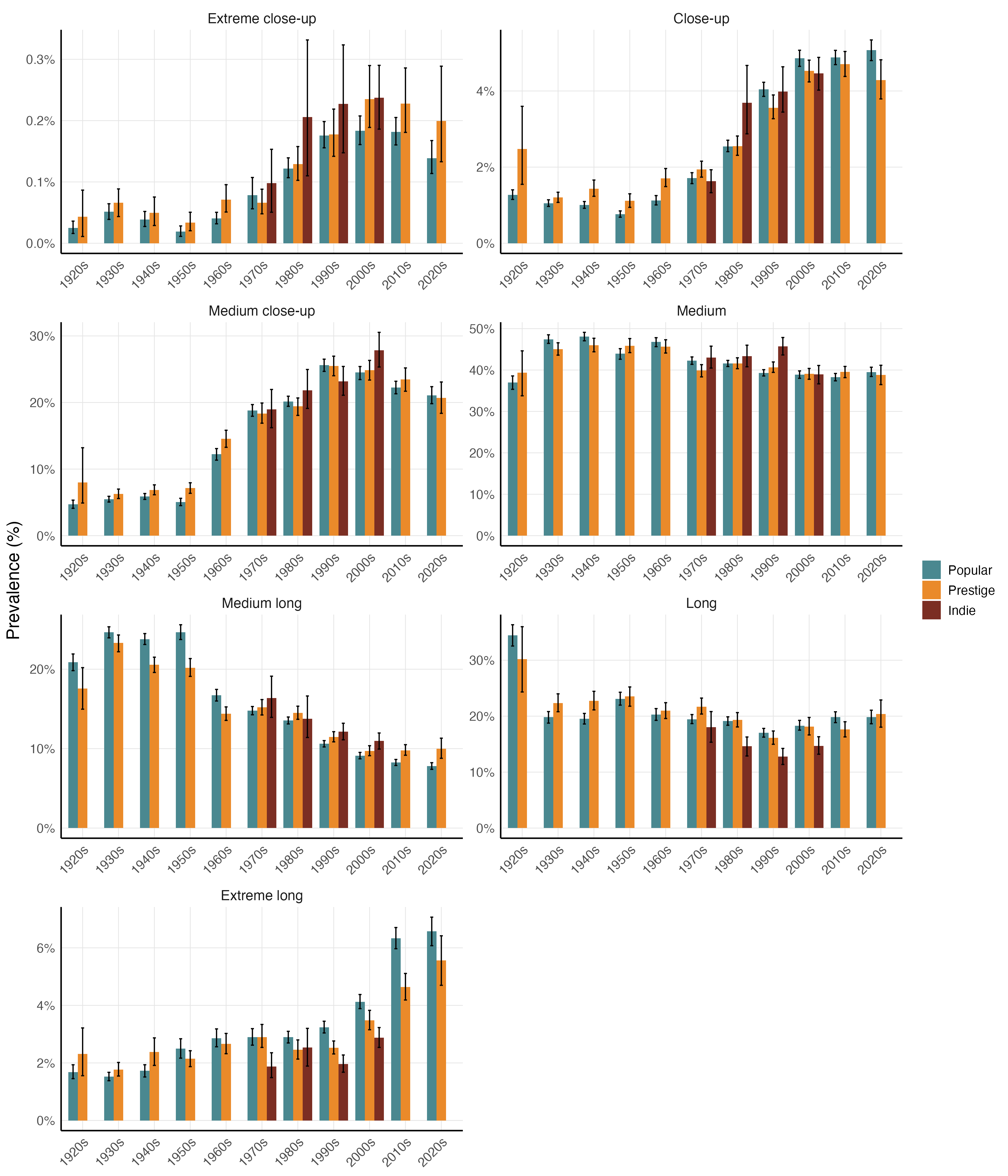}
    \caption{Prevalence of distance categories over time by collection (popular, prestige, American independent).}
    \label{fig:popprestigeindie}
\end{figure}

Figure \ref{fig:popprestigeindie} places all subcollections on the same chart in order to enable direct comparison between them. While independent films may have slightly less focus on longer shots, all subcollections generally display the same temporal dynamics: popular and prestige films both see a rise in the medium close-up between 1950-1975, and independent films occupy the same space when they arrive in the 1970s.

\FloatBarrier
\section{Gender differentials by category} 
\begin{figure}[htbp]
    \centering
    \includegraphics[width=1\linewidth]{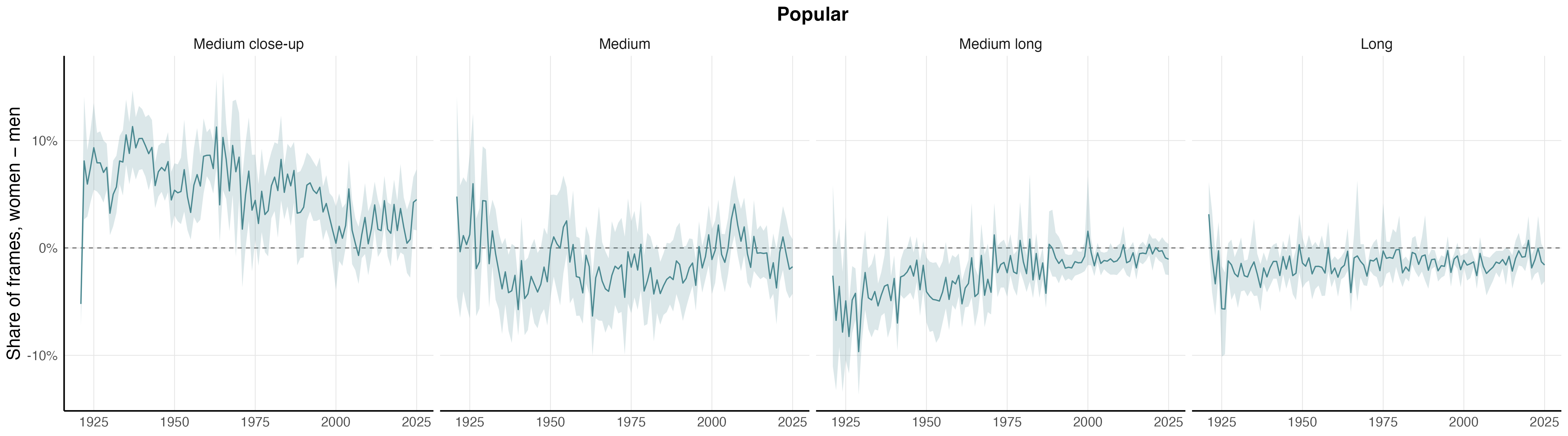}
    \caption{Difference in prevalence between shots focused on women vs. men over time, popular collection.}
    \label{fig:genderpoptime}
\end{figure}

\begin{figure}[htbp]
    \centering
    \includegraphics[width=1\linewidth]{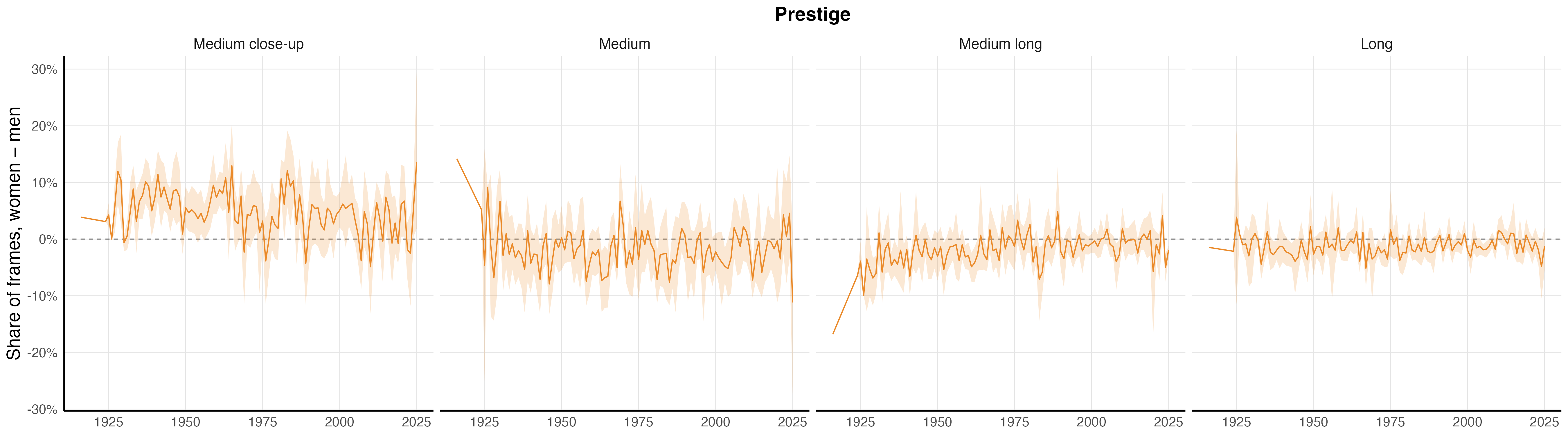}
    \caption{Difference in prevalence between shots focused on women vs. men over time, prestige collection.}
    \label{fig:genderprestigetime}
\end{figure}

\begin{figure}[htbp]
    \centering
    \includegraphics[width=1\linewidth]{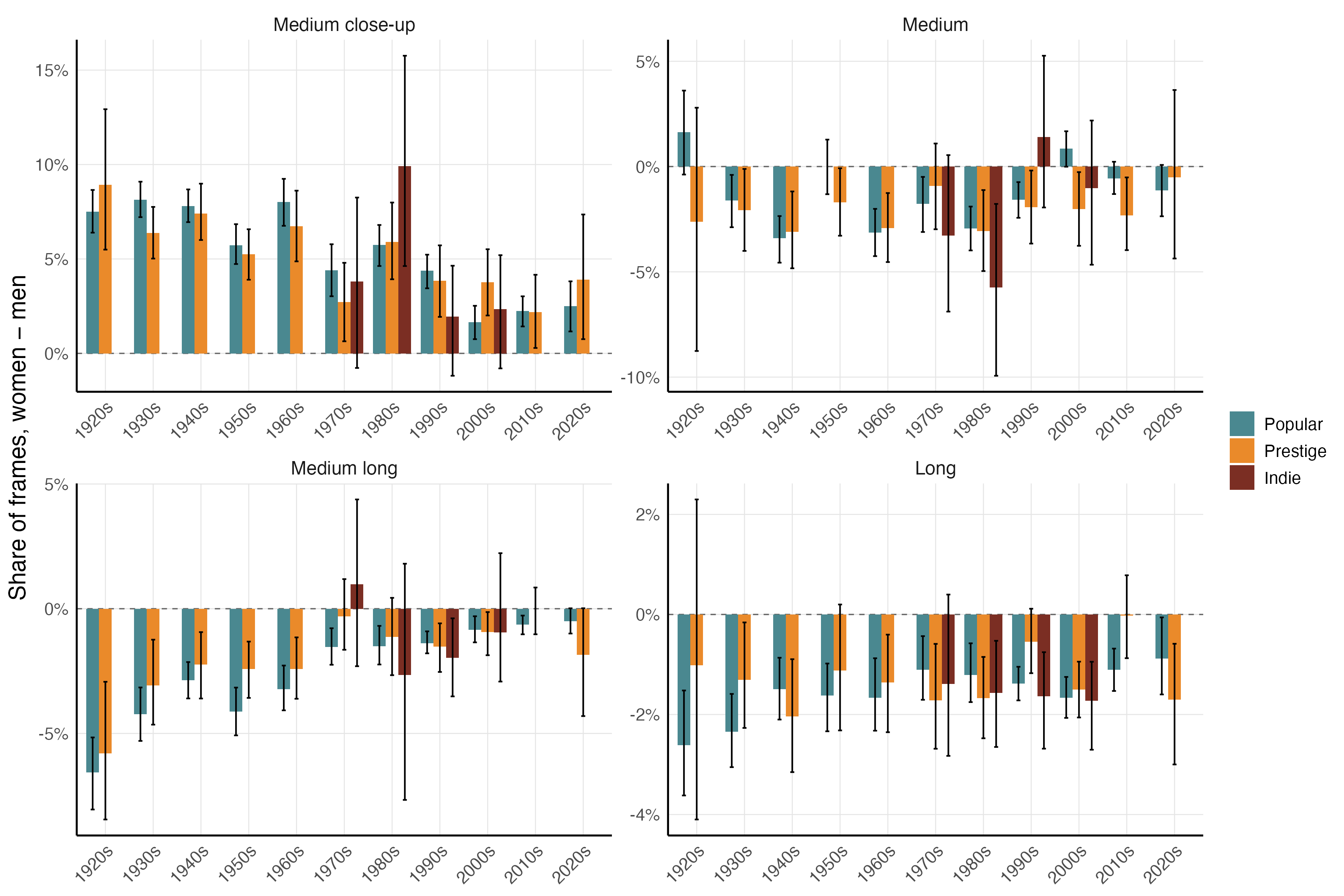}
    \caption{Difference in prevalence between shots focused on women vs. men over time, all collections.}
    \label{fig:genderalltime}
\end{figure}

Figures \ref{fig:genderpoptime}-\ref{fig:genderalltime} illustrate the gendered difference for each distance category. Categories above the 0\% y-axis line allocated comparatively more attention to women in shots of that distance.

\FloatBarrier
\section{Animated gender} Tables \ref{tab:gender-rate-paired-animated-human} and \ref{tab:gender-rate-paired-animated-non-human} list the prevalence of each distance category focused on a man (M) or woman (W), along with the difference between those rates. Table \ref{tab:gender-rate-paired-animated-human} presents results for human animated characters, and \ref{tab:gender-rate-paired-animated-non-human} does so for non-human characters.

\begin{table}[h!]
\centering
\begin{tabular}{lccc}
\toprule
Shot distance & M (\%) & W (\%) & W $-$ M (\%) \\
\midrule
Extreme close-up & 0.2 (0.1, 0.3) & 0.1 (0.1, 0.1) & $-$0.1 ($-$0.2, 0.0) \\
Close-up & 0.7 (0.5, 0.9) & 0.8 (0.7, 1.1) & 0.2 ($-$0.1, 0.4) \\
Medium close-up & 18.0 (16.2, 19.9) & 17.5 (15.6, 19.5) & $-$0.5 ($-$2.3, 1.5) \\
Medium & 48.6 (46.9, 50.2) & 49.0 (47.1, 50.9) & 0.4 ($-$1.5, 2.6) \\
Medium long & 9.0 (8.1, 9.9) & 10.3 (9.3, 11.4) & 1.2 (0.2, 2.4) \\
Long & 23.1 (20.9, 25.4) & 21.8 (19.9, 23.9) & $-$1.3 ($-$3.6, 0.6) \\
Extreme long & 0.4 (0.3, 0.6) & 0.4 (0.3, 0.7) & $-$0.0 ($-$0.1, 0.1) \\
\bottomrule
\end{tabular}
\caption{Share of a character's on-screen frames at each shot distance, for men (M) and women (W) among animated human characters in the year-matched animated/live-action sample, with 95\% bootstrap confidence intervals (resampling at the level of movies).}
\label{tab:gender-rate-paired-animated-human}
\end{table}

\begin{table}[h!]
\centering
\begin{tabular}{lccc}
\toprule
Shot distance & M (\%) & W (\%) & W $-$ M (\%) \\
\midrule
Extreme close-up & 0.3 (0.2, 0.5) & 0.2 (0.1, 0.3) & $-$0.1 ($-$0.3, 0.1) \\
Close-up & 1.2 (0.9, 1.6) & 1.3 (0.8, 1.9) & 0.1 ($-$0.4, 0.7) \\
Medium close-up & 16.0 (14.0, 17.9) & 16.2 (13.6, 18.7) & 0.2 ($-$1.9, 2.4) \\
Medium & 32.0 (29.8, 34.3) & 32.5 (29.4, 35.9) & 0.5 ($-$2.1, 3.7) \\
Medium long & 7.9 (6.7, 9.0) & 7.4 (6.3, 8.7) & $-$0.4 ($-$1.8, 0.8) \\
Long & 41.5 (38.6, 44.9) & 40.8 (36.9, 45.0) & $-$0.8 ($-$3.9, 2.5) \\
Extreme long & 1.1 (0.7, 1.5) & 1.6 (0.9, 2.5) & 0.5 (0.0, 1.2) \\
\bottomrule
\end{tabular}
\caption{Share of a character's on-screen frames at each shot distance, for men (M) and women (W) among animated non-human (animal or other) characters in the year-matched sample, with 95\% bootstrap confidence intervals (resampling at the level of movies).}
\label{tab:gender-rate-paired-animated-non-human}
\end{table}

\clearpage
Tables \ref{tab:gender-rate-paired-live-action} and \ref{tab:gender-rate-popular} present analysis for live-action movies; table \ref{tab:gender-rate-paired-live-action} shows only those movies in the paired animated-live action sample, while table \ref{tab:gender-rate-popular} shows all popular movies.

\begin{table}[h!]
\centering
\begin{tabular}{lccc}
\toprule
Shot distance & M (\%) & W (\%) & W $-$ M (\%) \\
\midrule
Extreme close-up & 0.0 (0.0, 0.0) & 0.0 (0.0, 0.0) & 0.0 ($-$0.0, 0.0) \\
Close-up & 0.3 (0.2, 0.3) & 0.4 (0.2, 0.6) & 0.1 ($-$0.0, 0.3) \\
Medium close-up & 36.7 (34.4, 38.9) & 39.7 (37.3, 42.2) & 3.0 (1.7, 4.4) \\
Medium & 48.9 (47.1, 50.5) & 47.5 (45.6, 49.2) & $-$1.4 ($-$2.6, $-$0.2) \\
Medium long & 8.4 (7.7, 9.2) & 7.9 (7.1, 8.8) & $-$0.6 ($-$1.3, 0.2) \\
Long & 5.7 (5.2, 6.2) & 4.5 (4.0, 5.2) & $-$1.1 ($-$1.7, $-$0.5) \\
Extreme long & 0.1 (0.0, 0.1) & 0.0 (0.0, 0.1) & $-$0.0 ($-$0.0, 0.0) \\
\bottomrule
\end{tabular}
\caption{Share of a character's on-screen frames at each shot distance, for men (M) and women (W) among characters in the live-action half of the year-matched sample, with 95\% bootstrap confidence intervals (resampling at the level of movies).}
\label{tab:gender-rate-paired-live-action}
\end{table}

\begin{table}[h!]
\centering
\begin{tabular}{lccc}
\toprule
Shot distance & M (\%) & W (\%) & W $-$ M (\%) \\
\midrule
Extreme close-up & 0.0 (0.0, 0.0) & 0.0 (0.0, 0.0) & 0.0 ($-$0.0, 0.0) \\
Close-up & 0.2 (0.2, 0.3) & 0.3 (0.3, 0.3) & 0.0 (0.0, 0.1) \\
Medium close-up & 25.6 (25.0, 26.1) & 30.6 (30.0, 31.2) & 5.0 (4.7, 5.3) \\
Medium & 51.0 (50.6, 51.3) & 49.7 (49.2, 50.1) & $-$1.3 ($-$1.6, $-$1.0) \\
Medium long & 14.6 (14.3, 14.9) & 12.4 (12.0, 12.7) & $-$2.2 ($-$2.4, $-$1.9) \\
Long & 8.5 (8.3, 8.8) & 7.0 (6.7, 7.3) & $-$1.5 ($-$1.7, $-$1.3) \\
Extreme long & 0.1 (0.1, 0.1) & 0.1 (0.1, 0.1) & $-$0.0 ($-$0.1, $-$0.0) \\
\bottomrule
\end{tabular}
\caption{Share of a character's on-screen frames at each shot distance, for men (M) and women (W) among characters in popular films, across all years, with 95\% bootstrap confidence intervals (resampling at the level of movies).}
\label{tab:gender-rate-popular}
\end{table}

\begin{figure}[htbp]
    \centering
    \includegraphics[width=.9\linewidth]{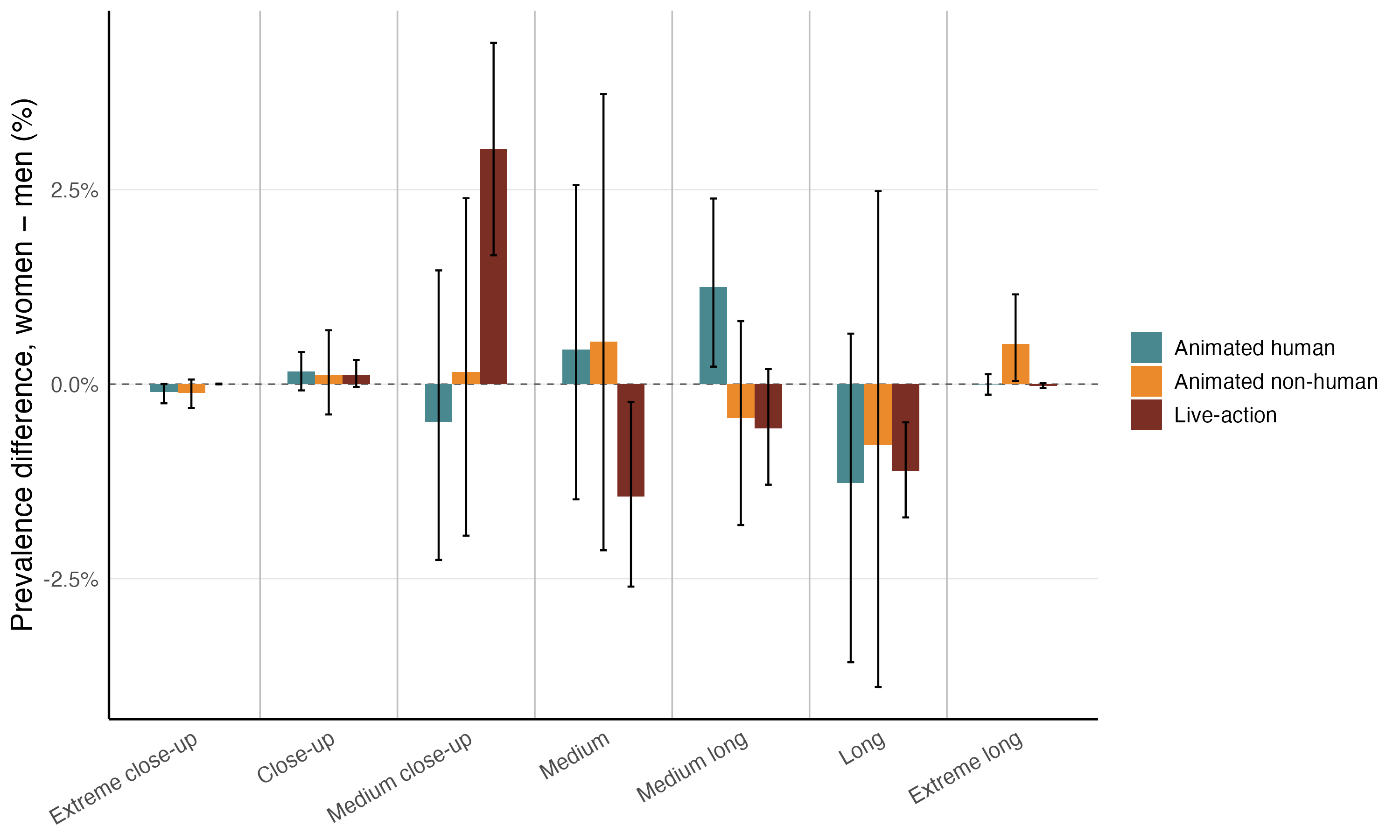}
    \caption{Paired test of animated (human, non-human) characters vs. live-action ones.}
    \label{fig:pairedanimation}
\end{figure}

\FloatBarrier
\clearpage
Figure \ref{fig:pairedanimation} places those results in comparison with live-action.

\FloatBarrier
\bibliography{bibliography}

\end{document}